\documentclass{IEEEtran}
\usepackage{amsmath,amssymb,amsfonts}
\usepackage{algorithmic}
\usepackage{graphicx}
\usepackage{textcomp}
\usepackage{booktabs}
\usepackage{ragged2e}
\usepackage{array}
\usepackage{multirow}
\usepackage{makecell}
\usepackage{wrapfig}
\usepackage[numbers]{natbib}
\usepackage{longtable} 
\usepackage{tabularx,colortbl}
\usepackage{threeparttable}
\usepackage[breaklinks]{hyperref}
\usepackage[utf8]{inputenc}
\usepackage{adjustbox}

\usepackage[caption=false,font=footnotesize]{subfig}
\usepackage{cuted}
\usepackage{capt-of}

\def\BibTeX{{\rm B\kern-.05em{\sc i\kern-.025em b}\kern-.08em
    T\kern-.1667em\lower.7ex\hbox{E}\kern-.125emX}}
\begin{document}
\title{Segment-Level Risk Discovery in Online Handwriting for Alzheimer’s Disease Detection
}
\author{Changqing Gong, Huafeng Qin and Mounîm A. El-Yacoubi

\thanks{Changqing Gong and Mounîm A. El-Yacoubi are with Telecom SudParis, Institut Polytechnique de Paris, 91120 Palaiseau, France (e-mail: changqing.gong@telecom-sudparis.eu; mounim.el\_yacoubi@telecom-sudparis.eu).}

\thanks{Huafeng Qin is with the School of Computer Science and Information Engineering, Chongqing Technology and Business University, Chongqing 400067, China (e-mail: qinhuafengfeng@163.com).}

\thanks{Manuscript received xx; revised xx; accepted xx.
This work was supported in part by xxx, and in part by xxx.We also would like to thank Telecom SudParis and Institut Polytechnique de Paris for their partial support (Corresponding author: xxxxxx)}
}

\maketitle
\begin{abstract}
Online handwriting provides a non-invasive and low-cost behavioral biomarker for Alzheimer's disease (AD) detection, as it reflects both cognitive planning and fine motor control. Existing handwriting-based AD detection methods usually rely on global trajectory features or whole-sample representations, which can be strongly affected by individual writing style, task-specific variation, and acquisition noise. In this paper, we propose NormPaST-Risk, a healthy-normative Paper-Air selective trajectory state-space risk network for interpretable AD detection from online handwriting. Instead of treating the entire trajectory as a single holistic representation, our method reformulates AD handwriting detection as local disease-relevant segment discovery. Specifically, a multi-scale temporal encoder captures stroke dynamics at different temporal resolutions, while a selective Paper-Air state-space encoder models long-range handwriting progression and distinguishes on-paper motor execution from in-air planning and transition behaviors. To explicitly characterize abnormal deviations, a healthy normative branch learns normal handwriting dynamics from healthy controls, and a task-aware multi-expert segment-risk module estimates segment-level AD risk calibrated by hidden-state changes and normative deviations. A weakly supervised segment-level objective further enables high-risk segment discovery without manual segment annotations. Experiments on the DARWIN benchmark demonstrate that the proposed framework achieves superior AD/HC classification performance compared with existing methods. Moreover, the discovered high-risk segments can be projected back to the original handwriting trajectory, providing interpretable evidence associated with AD-related handwriting variations.
\end{abstract}

% Main text
\section{Introduction}\label{Introduction}

Alzheimer's disease (AD), the most common cause of dementia, is a progressive neurodegenerative disorder characterized by cognitive decline affecting memory, reasoning, and daily functioning \cite{knopman2021alzheimer}. Early identification is increasingly important for timely clinical intervention and disease management \cite{long2019alzheimer(11),jack2024revised}. However, established biomarker modalities, such as amyloid positron emission tomography (PET) and cerebrospinal fluid (CSF) assays, remain constrained in large-scale screening by their cost, limited accessibility, and/or invasiveness \cite{hampel2018blood,karikari2022blood}. Neuropsychological assessments, such as the Mini-Mental State Examination (MMSE) and Montreal Cognitive Assessment (MoCA), are more accessible, but their performance can be influenced by cultural and educational factors, while repeated assessments may also be affected by practice effects and rater dependence \cite{tsoi2015cognitive,kourtis2019digital}. Therefore, there remains a strong need for accessible, objective, and low-cost approaches to support large-scale AD screening.

To this end, a variety of behavioral signals, including eye movements, speech, gait, and motor activity, have been investigated as potential digital markers of cognitive decline and dementia \cite{qi2025alzheimer,vcepukaityte2024early}. Among these behavioral modalities, handwriting is particularly promising because it requires the coordinated engagement of cognitive, visuospatial, linguistic, and fine motor processes \cite{yan2008alzheimer(21),fernandes2023handwriting}. With digital tablets, online handwriting tasks can capture fine-grained spatiotemporal dynamics, including pen trajectories, pressure, velocity, pauses, and in-air movements, providing an unobtrusive window into cognitive-motor function \cite{el2019aging(1),kawa2017spatial(17)}. Previous studies have reported alterations in handwriting kinematics and dynamics in individuals with AD and mild cognitive impairment, including changes in movement timing, fluency, pressure, and in-air behavior \cite{fernandes2023handwriting,el2019aging(1),kawa2017spatial(17),cilia2024word}.

% Handwriting-based AD detection has traditionally relied on manually engineered task-level kinematic and spatial features combined with conventional machine learning classifiers \cite{kahindo2018characterizing,cilia2022diagnosing}. More recent studies have shifted toward learned representations, including sequential modeling \cite{el2019aging(1)}, deep transfer learning from handwriting dynamics \cite{cilia2021online}, 1D convolutional modeling \cite{dao2022detection}, and hybrid architectures integrating 1D dynamic signals with 2D handwriting images \cite{gong2025hybrid}. More recently, Nardone et al. analyzed individual handwriting strokes, demonstrating the value of preserving fine-grained local dynamics rather than relying solely on task-level aggregation \cite{nardone2025handwriting}. Nevertheless, existing methods mainly operate at the task or predefined stroke level, leaving the disease relevance of finer local trajectory segments largely unexplored.

Handwriting-based AD detection has traditionally relied on manually engineered 
kinematic, spatial, and pressure-related features combined with conventional 
machine learning classifiers 
\cite{ghaderyan2018new,cilia2022diagnosing,garre2017kinematic(16)}. 
With the development of deep learning, recent studies have increasingly explored 
learned handwriting representations, including sequential representation learning 
\cite{el2019aging(1)}, convolutional neural networks 
\cite{erdogmus2023promise}, deep transfer learning 
\cite{cilia2021online}, and multimodal Transformer architectures integrating 
1D dynamic signals with 2D handwriting images \cite{gong2025hybrid}. 
Beyond task-level classification, recent research has also investigated more 
fine-grained handwriting characteristics, including the effects of word semantics 
and phonology on handwriting dynamics \cite{cilia2024word} and individual 
stroke-level representations for AD prediction \cite{nardone2025handwriting}. 
These developments are consistent with broader evidence that AD-related changes 
can manifest across motor, visuospatial, and linguistic aspects of handwriting 
\cite{fernandes2023handwriting}.

Building on these advances toward finer-grained handwriting analysis, an important question is whether disease-related handwriting evidence can be more effectively captured at the local trajectory level. Such evidence may not be uniformly distributed throughout an entire handwriting sequence, while global or task-level representations can be influenced by non-disease-related variations, including individual writing style, habitual pen usage, task-specific execution strategies, and acquisition noise. These factors may dilute or obscure subtle local abnormalities. Preserving local trajectory structure may therefore provide a more direct way to identify disease-relevant patterns while reducing the influence of irrelevant global variations. Similar observations have also been reported in recent time-series studies, where modeling informative local patches or discriminative subsequences has shown advantages over relying exclusively on global representations \cite{park2026paano,liu2026unified}.

Learning such localized disease-related handwriting evidence, however, is challenging because only a single label is available for each complete handwriting trajectory, while no segment-level annotations indicate which local regions contain disease-related abnormalities. \textcolor{blue}{We therefore formulate segment-risk learning in a weakly supervised manner. The model assigns a disease-risk score to each local segment and uses the segments with the highest predicted risks as local evidence under trajectory-level supervision.} This encourages the model to focus on a subset of informative handwriting regions that are more relevant to disease discrimination. Based on this motivation, we propose NormPaST-Risk, a segment-level trajectory modeling framework for interpretable handwriting-based AD detection. The model first partitions online handwriting trajectories into state-consistent local segments that preserve Paper/Air boundaries, and a multi-scale temporal encoder captures short- and mid-range kinematic dynamics. A Selective Paper-Air state-space encoder further models long-range dependencies while explicitly incorporating pen-contact states to distinguish on-paper motor execution from in-air planning and transition behavior. A healthy normative branch learns normal handwriting dynamics from healthy controls, while a task-aware multi-expert segment-risk module estimates local disease risks and identifies diagnostically informative segments. Together, these components integrate local abnormality discovery with long-range trajectory modeling for interpretable AD detection. The main contributions of this work are summarized as follows:

\begin{itemize}

\item \textbf{}
We introduce segment-level disease-risk modeling into online handwriting-based AD classification. Instead of treating all trajectory regions as equally informative, the proposed framework explicitly estimates segment-level disease relevance and uses the learned risks to adaptively weight local representations while retaining the complete trajectory context.

\item \textbf{}
We propose NormPaST-Risk, a novel trajectory model that integrates multi-scale temporal encoding, Paper-Air state modulation, selective state-space modeling, a healthy normative branch, and task-aware multi-expert segment-risk learning. The framework captures both local and long-range handwriting dynamics while enabling interpretable localization of disease-relevant segments.

\item \textbf{}
Extensive experiments demonstrate the effectiveness of NormPaST-Risk. The proposed method consistently outperforms representative machine learning and sequence-modeling baselines, while ablation and segment-deletion studies further validate the contributions of its key components and the discriminative value of the discovered high-risk handwriting segments.

\end{itemize}

The remainder of this paper is organized as follows: Section 2 reviews related work, Section 3 presents NormPaST-Risk, Section 4 reports the experiments, and Section 5 concludes the paper.

\section{Related Work}
\label{sec:related}

Dynamic handwriting analysis has been widely investigated as a non-invasive means of characterizing cognitive and motor alterations associated with neurodegenerative disorders \cite{impedovo2018dynamic,fernandes2023handwriting}. Early handwriting-based AD detection mainly relied on manually engineered kinematic and spatial descriptors, such as writing duration, in-air and on-paper time, velocity, acceleration, pressure, jerk, and trajectory statistics, followed by conventional machine learning classifiers \cite{ghaderyan2018new,cilia2022diagnosing}. The release of the DARWIN dataset further enabled systematic evaluation across multiple handwriting and drawing tasks \cite{cilia2022diagnosing}. Although these handcrafted representations are interpretable, they typically summarize an entire task and may therefore obscure subtle local abnormalities.

More recent studies have shifted toward learned representations of handwriting dynamics. El-Yacoubi et al. introduced sequential representation learning to capture temporal handwriting behavior beyond global kinematic statistics \cite{el2019aging(1)}. Cilia et al. encoded online handwriting dynamics into synthetic images for deep transfer learning \cite{cilia2021online}, while Dao et al. directly modeled handwriting sequences using a 1D convolutional neural network \cite{dao2022detection}. Erdogmus and Kabakus further explored convolutional representations derived from handwriting signals \cite{erdogmus2023promise}. More recently, attention-based and multimodal architectures have been investigated: Kang et al. employed self-attention across multiple handwriting tasks \cite{kang2024early}, whereas Gong et al. integrated 1D dynamic signals with 2D handwriting images using a hybrid Transformer \cite{gong2025hybrid}. Despite increasingly powerful representation learning, these methods primarily focus on task- or sample-level prediction.

Fine-grained handwriting dynamics have consequently received increasing attention. Online handwriting contains both on-paper and in-air movements, whose temporal and kinematic characteristics can provide complementary information about handwriting behavior \cite{fernandes2023handwriting,kawa2017spatial(17)}. Cilia et al. further investigated how linguistic factors such as word semantics and phonology affect handwriting dynamics in AD \cite{cilia2024word}. More importantly, Nardone et al. analyzed individual handwriting strokes rather than compressing each task into a single aggregated feature vector, demonstrating the benefit of preserving fine-grained movement information \cite{nardone2025handwriting}. Explainability studies have also attempted to identify discriminative temporal patterns in online handwriting; for example, Sweidan et al. interpreted CNN predictions and revealed localized movement characteristics associated with AD \cite{sweidan2024explainability}. Nevertheless, existing fine-grained approaches mainly rely on predefined stroke units or post-hoc interpretation rather than directly learning disease relevance for local trajectory segments.

Online handwriting is also inherently sequential, requiring the modeling of both short-term motor variations and long-range writing dependencies. State-space models (SSMs) provide an efficient framework for long-sequence modeling. Structured and diagonal SSMs enable efficient modeling of long-range dependencies \cite{gu2021efficiently,gupta2022diagonal,gu2022parameterization}, while selective SSMs introduce input-dependent state propagation to adapt sequence dynamics to the current input \cite{gu2023mamba}. However, SSM-based modeling remains largely unexplored for handwriting-based AD detection. In particular, existing handwriting approaches typically summarize Paper/Air behavior through temporal or kinematic descriptors rather than explicitly using pen-contact state to modulate long-range sequence dynamics.

Overall, handwriting-based AD detection has progressed from handcrafted task-level descriptors to deep sequential, attention-based, multimodal, and stroke-level representations. However, three limitations remain. First, most methods perform prediction using global task representations or predefined stroke units rather than explicitly learning disease-relevant local segments from trajectory-level supervision. Second, Paper/Air information is generally represented through derived features rather than explicitly modeled as contextual state governing trajectory dynamics. Third, localized disease-risk discovery, long-range sequence modeling, and deviations from healthy handwriting dynamics have not yet been jointly modeled within a unified framework. These limitations motivate NormPaST-Risk, which constructs state-consistent Paper-Air segments, models long-range trajectory dynamics using a selective state-space encoder, learns healthy normative dynamics, and discovers high-risk local handwriting segments under weak supervision.

\section{Proposed Approach}
\label{sec:proposed_approach}

Handwriting-based Alzheimer's disease (AD) detection is challenging because
disease-related abnormalities may be sparse, localized, and task-dependent.
A complete online handwriting trajectory contains substantial variations
arising from individual writing style, task-specific execution, and acquisition
noise, whereas discriminative AD-related evidence may appear only in short local
segments, such as hesitation, unstable stroke execution, abnormal in-air
movement, or irregular Paper-Air transitions.
Therefore, instead of representing a handwriting sample solely by a global
trajectory descriptor, we propose NormPaST-Risk, a healthy-normative
Paper-Air selective trajectory state-space risk network for interpretable
AD detection from online handwriting.

\begin{figure*}[!t]
    \centering
    \includegraphics[
        width=\textwidth,
        height=\textheight,
        keepaspectratio
    ]{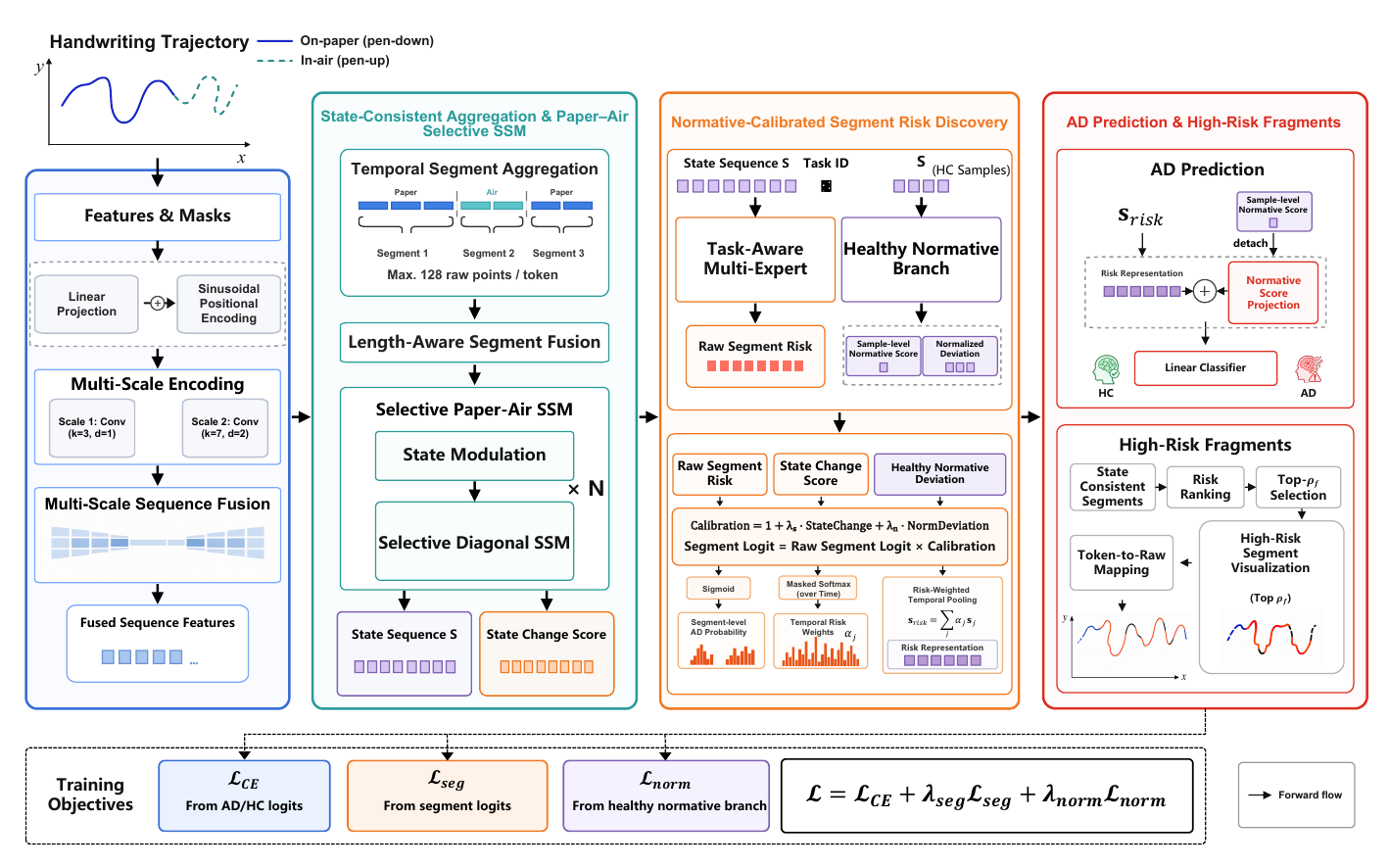}
    \caption{
    Overall architecture of NormPaST-Risk.
    The framework consists of online handwriting representation,
    multi-scale trajectory encoding, state-aware segmentation and
    Paper-Air selective state-space modeling,
    normative-calibrated segment-risk discovery,
    and AD prediction.
    }
    \label{fig:framework}
\end{figure*}

The proposed framework is motivated by four considerations.
First, online handwriting is a long temporal sequence with multi-scale dynamics,
requiring both short-range stroke dynamics and long-range writing progression
to be modeled.
Second, on-paper and in-air movements convey different behavioral information:
Paper trajectories primarily reflect executed motor behavior, whereas Air
trajectories may encode planning, hesitation, and cognitive-motor transitions.
Third, local AD-related abnormalities can be more explicitly characterized by
measuring deviations from healthy handwriting dynamics.
Fourth, AD-related manifestations may vary across handwriting tasks and
local trajectory segments, motivating task-aware multi-expert risk modeling
to capture heterogeneous abnormal patterns and localize disease-relevant
handwriting evidence. As shown in Fig.\ref{fig:framework}, NormPaST-Risk consists of six components:
(1) a multi-scale encoder,
(2) a temporal segment aggregator,
(3) a Selective Paper-Air SSM,
(4) a healthy normative branch,
(5) a task-aware multi-expert segment-risk module, and
(6) a classification head with weakly supervised segment-risk learning.

\subsection{Input Representation}
\label{subsec:input_representation}

Given an online handwriting sample, we denote the raw trajectory as
\begin{equation}
    \mathbf{X}
    =
    \{\mathbf{x}_t\}_{t=1}^{T_r},
    \quad
    \mathbf{x}_t \in \mathbb{R}^{F},
\end{equation}
where $T_r$ is the raw sequence length and $F$ is the feature dimension. 
We define the raw valid mask, paper mask, and air mask as
$\mathbf{m},\mathbf{m}^{p},\mathbf{m}^{a}
    \in \{0,1\}^{T_r}$,
where $m_t=1$ indicates a valid raw point, $m_t^{p}=1$ indicates a pen-down point, and $m_t^{a}=1$ indicates a pen-up point. 
For every valid point, the paper and air states are mutually exclusive.

% We also extract a transition sequence around paper-air state changes:
% \begin{equation}
%     \mathbf{U}
%     =
%     \{\mathbf{u}_n\}_{n=1}^{N},
%     \quad
%     \mathbf{u}_n \in \mathbb{R}^{F_{tr}},
% \end{equation}
% where $N$ is the number of valid transition events. 
% Each transition descriptor includes the previous pen state, next pen state, local position, pressure before and after transition, and local air/paper duration statistics.

\subsection{Multi-Scale Trajectory Encoder}
\label{subsec:multi_scale_encoder}

AD-related handwriting abnormalities may occur at different temporal scales.
For example, short receptive fields can capture local stroke instability, while longer receptive fields can capture hesitation and movement organization.
Therefore, we first encode the raw trajectory with a multi-scale temporal encoder.

Each raw point is projected into a latent space and augmented with positional encoding:
\begin{equation}
    \mathbf{f}_{t}
    =
    \phi_{\mathrm{in}}(\mathbf{x}_t)
    +
    \mathbf{p}_t,
\end{equation}
where $\phi_{\mathrm{in}}(\cdot)$ is a linear projection and $\mathbf{p}_t$ is the positional encoding. Then, $Q$ temporal convolutional streams with different kernel sizes and dilation rates are applied:
\begin{equation}
    \mathbf{F}^{q}
    =
    \mathcal{E}_{q}(\mathbf{f}_{t}),
    \quad
    q=1,\ldots,Q,
\end{equation}
where $\mathbf{F}^{q}=\{\mathbf{f}^{q}_{t}\}_{t=1}^{T_r}$ and $\mathcal{E}_{q}(\cdot)$ denotes the $q$-th temporal stream. 
Each stream preserves the original sequence length.

To adaptively fuse different scales at each raw point, we compute scale weights:
\begin{equation}
    a_{t,q}
    =
    \frac{
        \exp\left(s(\mathbf{f}^{q}_{t})\right)
    }{
        \sum_{q=1}^{Q}
        \exp\left(s(\mathbf{f}^{q}_{t})\right)
    },
\end{equation}
where $s(\cdot)$ is a learnable scoring function. 
The fused point-wise representation is
\begin{equation}
    \bar{\mathbf{f}}_{t}
    =
    \sum_{q=1}^{Q}
    a_{t,q}\mathbf{f}^{q}_{t}.
\end{equation}

This operation produces a point-wise sequence
\begin{equation}
    \bar{\mathbf{F}}
    =
    \{\bar{\mathbf{f}}_t\}_{t=1}^{T_r},
    \quad
    \bar{\mathbf{f}}_t \in \mathbb{R}^{D}.
\end{equation}

\subsection{Temporal Segment Aggregation}
\label{subsec:temporal_segment_aggregation}

Directly feeding all points into the sequence encoder can be inefficient and sensitive to point-level noise. 
However, naive fixed-stride downsampling may merge pen-down and pen-up states into the same token, which weakens the meaning of paper-air modeling. 
To address this issue, we introduce a state-consistent temporal segment aggregator.

\textcolor{blue}{Let $\gamma$ denote the maximum number of trajectory points contained in each local segment. The aggregator partitions the valid trajectory into a sequence of segment intervals:}
\begin{equation}
    \mathcal{I}
    =
    \{I_j\}_{j=1}^{T_f},
    \quad
    I_j=[s_j,e_j),
\end{equation}
where $T_f$ denotes the number of local segments. Each interval satisfies
\begin{equation}
    1
    \leq
    \ell_j
    =
    e_j-s_j
    \leq
    \gamma.
\end{equation}
A Paper/Air state transition always terminates the current interval. 
Therefore, all points inside the same interval share the same pen state.

\textcolor{blue}{For each segment interval $I_j=[s_j,e_j)$, we first aggregate the encoded point-wise trajectory representations within the interval by mean pooling:}

\begin{equation}
    \boldsymbol{\mu}_j
    =
    \frac{1}{\ell_j}
    \sum_{t=s_j}^{e_j-1}
    \bar{\mathbf{f}}_t ,
\end{equation}

The scale weights and the valid, paper, and air masks are aggregated according to the same segment intervals. Since the intervals are constructed without crossing Paper/Air boundaries, each segment-level representation is aligned with a specific raw-point interval and belongs to a single pen-contact state.

\subsection{Length-Aware Segment Token Fusion}
\label{subsec:length_aware_segment_fusion}

Mean pooling provides a stable gradient path for all points inside a segment, but it can dilute salient local responses. 
Therefore, we additionally compute a channel-wise max-pooled token:
\begin{equation}
    \boldsymbol{\nu}_j
    =
    \max_{t\in I_j}
    \bar{\mathbf{f}}_t,
\end{equation}
where the maximum is taken independently for each channel.

The mean token and max token are combined by a learnable channel-wise convex mixture:
\begin{equation}
    \mathbf{f}_j
    =
    \boldsymbol{\mu}_j
    +
    \mathbf{w}
    \odot
    \left(
    \boldsymbol{\nu}_j-\boldsymbol{\mu}_j
    \right),
    \quad
    \mathbf{w}
    =
    \sigma(\boldsymbol{\theta}),
\end{equation}
where $\boldsymbol{\theta}\in\mathbb{R}^{D}$ is a learnable parameter, $\mathbf{w}\in(0,1)^{D}$, and $\odot$ denotes element-wise multiplication. 
This design keeps the stable mean-pooling path while giving salient responses an additional length-independent path.

Because segments can have different true lengths, especially when a segment is terminated early by a Paper/Air transition, we encode the duration of each segment using two descriptors:
\begin{equation}
    \ell_j^{\mathrm{lin}}
    =
    \frac{\ell_j}{\gamma},
    \qquad
    \ell_j^{\mathrm{log}}
    =
    \frac{\log(1+\ell_j)}
    {\log(1+\gamma)}.
\end{equation}

The final segment token is
\begin{equation}
    \mathbf{z}_j
    =
    \mathbf{f}_j
    +
    \phi_{\ell}
    ([\ell_j^{\mathrm{lin}},
        \ell_j^{\mathrm{log}}]),
\end{equation}
where $\phi_{\ell}(\cdot)$ is a learnable duration projection. 
The resulting segment-token sequence is
\begin{equation}
    \mathbf{Z}
    =
    \{\mathbf{z}_j\}_{j=1}^{T_f},
    \quad
    \mathbf{z}_j\in\mathbb{R}^{D}.
\end{equation}

\begin{figure}[ht]
\centerline{\includegraphics[width=0.85\columnwidth]{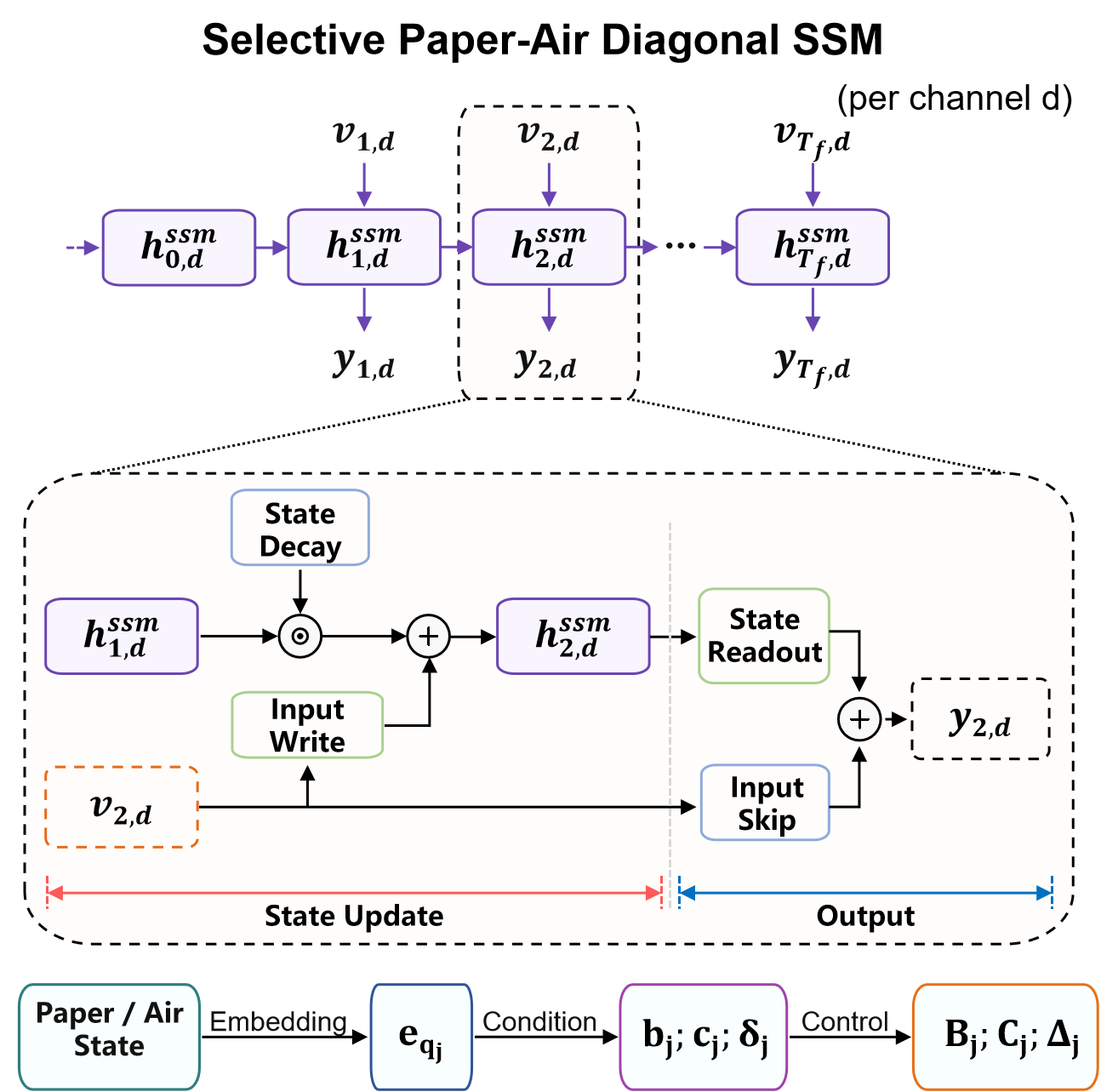}}
\caption{Illustration of the selective diagonal SSM.}
\label{fig:ssm}
\end{figure}

\subsection{Selective Paper-Air State-Space Encoder}
\label{subsec:selective_paper_air_ssm}

Dynamic handwriting inherently couples motor execution with higher-level cognitive processes such as planning and coordination, while on-paper and in-air movements exhibit distinct behavioral characteristics and have shown different sensitivity to cognitive impairment \cite{impedovo2018dynamic,muller2017increased}.
This motivates us to treat the observed Paper/Air state as contextual information rather than as an ordinary kinematic feature.
Inspired by feature-wise conditional modulation \cite{perez2018film}, we use layer-specific Paper-Air modulation to adapt intermediate trajectory representations to the observed pen state at different abstraction levels.
The resulting state-aware sequence is subsequently modeled by selective state-space dynamics.
Structured and diagonal SSMs have demonstrated strong capability for long-range sequence modeling \cite{gu2021efficiently,gupta2022diagonal,gu2022parameterization}, while selective SSMs further enable input-dependent state propagation through dynamically parameterized state-space components \cite{gu2023mamba}.
These observations motivate a hierarchical encoder in which Paper/Air state modulates the representation space and selective SSMs adapt the temporal state evolution.

The segment-token sequence still contains long-range temporal dependencies.
To model them efficiently, we use a Selective Paper-Air state-space encoder composed of $L$ stacked selective SSM blocks.
At each layer $l$, the current token representation is adaptively modulated according to its observed Paper/Air state:
\begin{equation}
\tilde{\mathbf{h}}^{l}_j
=
\left(
\mathbf{h}^{l-1}_j
+
\mathbf{e}^{l}_{q_j}
\right)
\odot
\left(
0.75
+
0.5\mathbf{g}^{l}_j
\right),
\end{equation}
where $\mathbf{h}^{(0)}_j=\mathbf{z}_j$, $q_j$ denotes the observed Paper/Air state, $\mathbf{e}^{l}_{q_j}\in\mathbb{R}^{D}$ is its layer-specific learnable latent embedding, and $\mathbf{g}^{l}_j\in(0,1)^D$ is a layer-specific Paper-Air modulation gate predicted from the concatenation of $\mathbf{h}^{l-1}_j$ and $\mathbf{e}^{l}_{q_j}$.
Although the observed state $q_j$ remains unchanged across layers, its latent representation and modulation effect are learned independently at different representation levels.

Inside the $l$-th selective diagonal SSM block, a gated input token is computed as
\begin{equation}
\mathbf{v}^{l}_j
=
\mathbf{o}^{l}_j
\odot
\sigma(\mathbf{g}^{s,l}_j),
\quad
[\mathbf{o}^{l}_j;\mathbf{g}^{s,l}_j]
=
\phi_{\mathrm{in}}^{s,l}
\left(
\mathrm{LN}(\tilde{\mathbf{h}}^{l}_j)
\right).
\end{equation}
where $\phi_{\mathrm{in}}^{s,l}$ denotes a learnable linear input projection in the $l$-th selective SSM layer. 

The current token also generates channel-wise selection factors:
\begin{equation}
[\mathbf{b}^{l}_j;
\mathbf{c}^{l}_j;
\boldsymbol{\delta}^{l}_j]
=
\phi_{\mathrm{sel}}^{l}
\left(
\mathrm{LN}(\tilde{\mathbf{h}}^{l}_j)
\right),
\end{equation}
where
$\mathbf{b}^{l}_j,
\mathbf{c}^{l}_j,
\boldsymbol{\delta}^{l}_j
\in\mathbb{R}^{D}$, $\phi_{\mathrm{sel}}^{l}$ denotes a learnable linear projection in the $l$-th selective SSM layer. 
The selection factors are transformed as
\begin{equation}
\mathbf{b}^{+,l}_j
=
2\sigma(\mathbf{b}^{l}_j),
\quad
\mathbf{c}^{+,l}_j
=
2\sigma(\mathbf{c}^{l}_j),
\end{equation}
\begin{equation}
\boldsymbol{\Delta}^{l}_j
=
\operatorname{Softplus}
\left(
\boldsymbol{\eta}^{l}_{\Delta}
+
0.5\tanh(\boldsymbol{\delta}^{l}_j)
\right),
\end{equation}
where $\boldsymbol{\eta}^{l}_{\Delta}\in\mathbb{R}^{D}$ is a learnable
channel-wise base parameter for the SSM step size.
The diagonal transition parameter is constrained to be stable:
\begin{equation}
\mathbf{A}^{l}
=
-\exp(\mathbf{A}^{l}_{\log}).
\end{equation}

For channel $d$, the SSM state is
$\mathbf{h}^{\mathrm{ssm},l}_{j,d}\in\mathbb{R}^{N_s}$,
where $N_s$ is the state dimension.
The selective diagonal recurrence is
\begin{equation}
\mathbf{h}^{\mathrm{ssm},l}_{j,d}
=
\exp
\left(
\mathbf{A}^{l}_{d}
\Delta^{l}_{j,d}
\right)
\odot
\mathbf{h}^{\mathrm{ssm},l}_{j-1,d}
+
\left(
b^{+,l}_{j,d}\mathbf{B}^{l}_{d}
\right)
v^{l}_{j,d},
\end{equation}
\begin{equation}
y^{l}_{j,d}
=
\left(
c^{+,l}_{j,d}\mathbf{C}^{l}_{d}
\right)^{\top}
\mathbf{h}^{\mathrm{ssm},l}_{j,d}
+
D^{l}_d v^{l}_{j,d},
\end{equation}
where
$\mathbf{A}^{l}_{d},
\mathbf{B}^{l}_{d},
\mathbf{C}^{l}_{d}\in\mathbb{R}^{N_s}$
are channel-wise SSM parameters, and $D^{l}_d$ is a learnable direct feedthrough parameter.

Finally, the channel-wise SSM responses are aggregated and projected to obtain
the SSM output $\mathbf{y}^{l}_j\in\mathbb{R}^{D}$.
The resulting output is then combined with the input of the current layer
through a residual connection followed by layer normalization.
The operations of the $l$-th Selective Paper-Air diagonal SSM layer can be
summarized as
\begin{equation}
\mathbf{h}^{l}_j
=
\mathrm{LN}
\left(
\mathbf{h}^{l-1}_j
+
\mathcal{D}_{\mathrm{SSM}}^{l}
\left(
\mathcal{M}_{\mathrm{PA}}^{l}
\left(
\mathbf{h}^{l-1}_j,q_j
\right)
\right)
\right).
\end{equation}
Here, $\mathcal{M}_{\mathrm{PA}}^{l}$ represents the Paper-Air modulation, while $\mathcal{D}_{\mathrm{SSM}}^{l}$ represents the selective diagona SSM transformation, including state propagation, input writing, state readout, and output projection.

After the final layer, we obtain the context-aware state sequence
\begin{equation}
    \mathbf{S}
    =
    \{\mathbf{s}_j\}_{j=1}^{T_f},
    \qquad
    \mathbf{s}_j
    =
    \mathbf{h}^{L}_j
    \in\mathbb{R}^{D}.
\end{equation}

To capture abrupt hidden-state changes, we compute a normalized state-change score:
\begin{equation}
    \chi_j =
    \frac{
        \left\|
        \mathbf{s}_j-\mathbf{s}_{j-1}
        \right\|_2
    }{
        \max\left(
        \max_{k}
        \left\|
        \mathbf{s}_k-\mathbf{s}_{k-1}
        \right\|_2,
        \epsilon
        \right)
    },
    \qquad j \ge 2,
\end{equation}
This score is used only as a calibration factor for learned segment risk, rather than as an independent disease indicator.

\subsection{Healthy Normative Branch}
\label{subsec:healthy_normative_branch}

\begin{figure}[ht]
\centerline{\includegraphics[width=0.85\columnwidth]{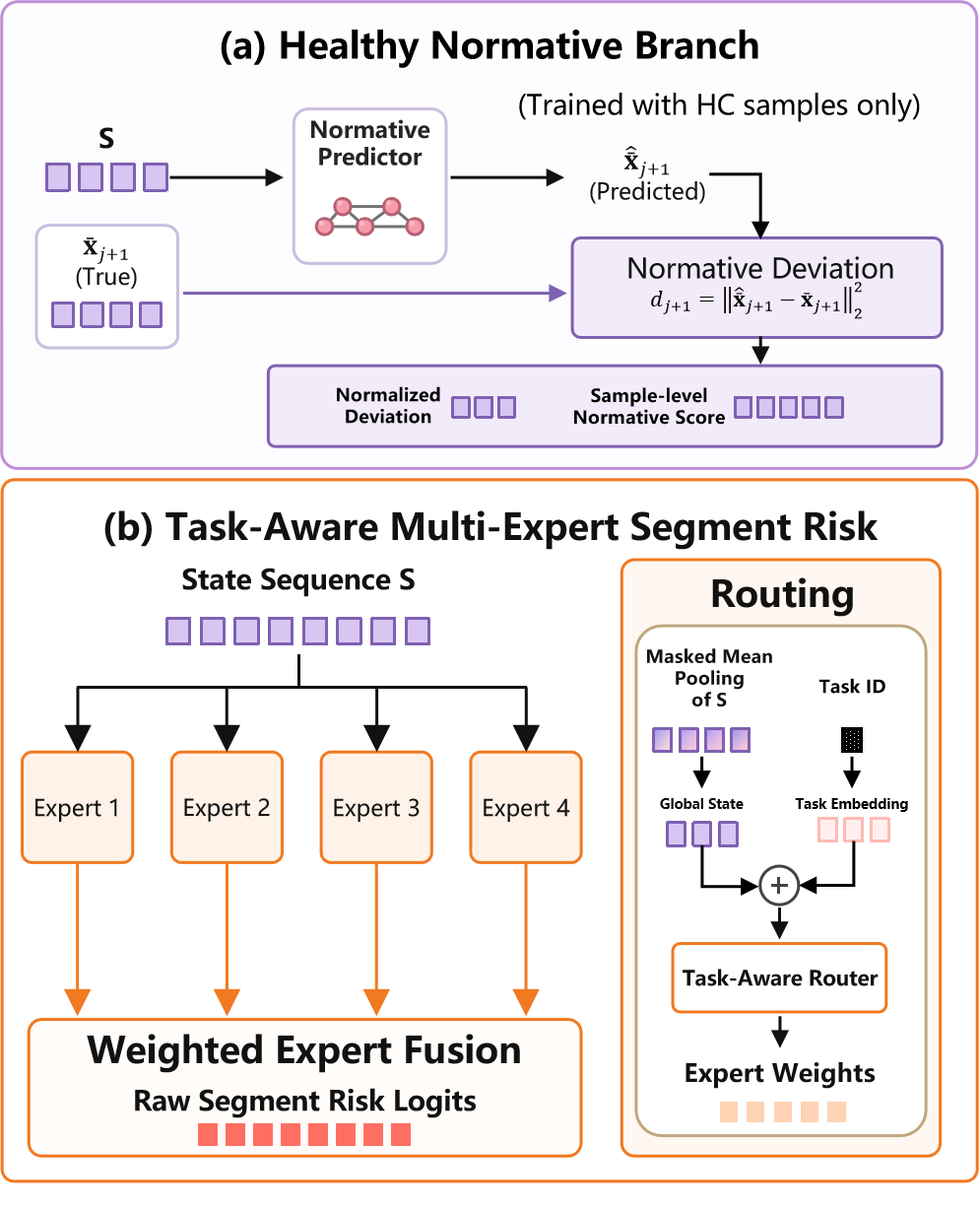}}
\caption{Illustration of (a) the Healthy Normative Branch and (b) the Task-Aware Multi-Expert Segment Risk module.}
\label{fig:Segment_Risk_Healthy_Normative}
\end{figure}

To explicitly characterize healthy handwriting patterns, we introduce a
healthy normative branch, as shown in Fig.\ref{fig:Segment_Risk_Healthy_Normative} (a). Given the contextual state sequence
$\mathbf{S}$, the branch estimates the subsequent aggregated trajectory as
\begin{equation}
    \hat{\bar{\mathbf{x}}}_{j+1}
    =
    f_{\mathrm{norm}}(\mathbf{s}_j),
    \quad
    j=1,\ldots,T_f-1.
\end{equation}
Here, $\mathbf{s}_j$ is the context-aware state representation of the $j$-th segment, and $\bar{\mathbf{x}}_j$ denotes the mean-aggregated raw trajectory feature of the $j$-th segment. To ensure that $f_{\mathrm{norm}}$ captures healthy handwriting dynamics, its prediction objective is optimized exclusively using HC samples during training.

For valid consecutive segments, the normative deviation is computed as
\begin{equation}
    d_{j+1}
    =
    \frac{1}{F}
    \left\|
    \hat{\bar{\mathbf{x}}}_{j+1}
    -
    \bar{\mathbf{x}}_{j+1}
    \right\|_2^2,
\end{equation}
where $F$ denotes the dimensionality of the trajectory feature vector.
To align the deviation sequence with the segment-token sequence, we set
$d_1=0$.

The segment-wise deviation is normalized within each sample as
\begin{equation}
    \tilde{d}_{j}
    =
    \frac{
        d_j
    }{
        \max
        \left(
            \max d_k,
            \epsilon
        \right)
    },
\end{equation}
where $\epsilon>0$ ensures numerical stability.

The sample-level normative score is computed from the unnormalized
deviation:
\begin{equation}
    d_{\mathrm{score}}
    =
    \frac{
        \sum_{j=1}^{T_f-1}
        m^{\mathrm{step}}_j d_{j+1}
    }{
        \sum_{j=1}^{T_f-1}
        m^{\mathrm{step}}_j
    },
\end{equation}
where $m^{\mathrm{step}}_j=1$ means that both tokens $j$ and $j+1$ are valid.

The normative prediction objective is optimized only using healthy
control samples. Therefore, the normative predictor learns
segment-level trajectory regularities characteristic of healthy
handwriting without directly fitting AD samples. During inference,
deviations from these healthy regularities provide auxiliary abnormality
cues for segment-risk calibration and subject-level classification.

\subsection{Task-Aware Multi-Expert Segment-Risk Modeling}
\label{subsec:multi_expert_segment_risk}

Different handwriting tasks may emphasize different cognitive and motor functions. 
Therefore, we use a task-aware multi-expert module to estimate token-level AD risk.

As shown in Fig.\ref{fig:Segment_Risk_Healthy_Normative} (b), each expert produces a segment-risk logit:
\begin{equation}
    r^{e}_{j}
    =
    f_e(\mathbf{s}_j),
    \quad
    e=1,\ldots,E,
\end{equation}
where $E$ is the number of experts. 
A global state vector is obtained by masked mean pooling:
\begin{equation}
    \mathbf{s}_{g}
    =
    \frac{
        \sum_{j=1}^{T_f}
        \bar{m}_j\mathbf{s}_j
    }{
        \sum_{j=1}^{T_f}
        \bar{m}_j
    },
\end{equation}
where $\bar{m}_j$ is the segment-level valid mask. 
The router input is the concatenation of the global state and the task embedding:
\begin{equation}
    \mathbf{z}_{r}
    =
    [\mathbf{s}_{g};\mathbf{e}_{\tau}],
\end{equation}
where $\tau$ denotes the task identity. 
The router produces expert logits:
\begin{equation}
    \boldsymbol{\rho}
    =
    \frac{
        R(\mathbf{z}_{r})
    }{
        \tau_r
    },
\end{equation}
where $\tau_r$ is the router temperature. The router logits are converted into normalized expert weights using
a softmax function:
\begin{equation}
    \pi_e
    =
    \frac{
        \exp(\rho_e)
    }{
        \sum_{l=1}^{E}
        \exp(\rho_l)
    },
    \quad
    e=1,\ldots,E.
\end{equation}

The raw segment-risk logit is
\begin{equation}
    r_j
    =
    \sum_{e=1}^{E}
    \pi_e r^{e}_{j}.
\end{equation}

The final segment-risk logit is calibrated by hidden-state change and healthy-normative deviation:
\begin{equation}
    \tilde{r}_j
    =
    r_j
    \left(
    1
    +
    \lambda_c\chi_j
    +
    \lambda_n\tilde{d}_j
    \right),
\end{equation}
where $\lambda_c$ and $\lambda_n$ control the contributions of state-change and normative-deviation calibration, respectively. 
The normative deviation is detached in this calibration path, so that AD labels do not directly optimize the healthy-only normative predictor through the segment-risk objective.

The segment-level AD probability is
\begin{equation}
    p_j
    =
    \sigma(\tilde{r}_j).
\end{equation}
The temporal risk weights are computed by masked softmax:
\begin{equation}
    \alpha_j
    =
    \frac{
        \bar{m}_j
        \exp(\tilde{r}_j)
    }{
        \sum_{k=1}^{T_f}
        \bar{m}_k
        \exp(\tilde{r}_k)
    }.
\end{equation}
The risk representation is obtained by
risk-weighted pooling:
\begin{equation}
    \mathbf{s}_{\mathrm{risk}}
    =
    \sum_{j=1}^{T_f}
    \alpha_j \mathbf{s}_j.
\end{equation}

\subsection{Final Classification}
\label{subsec:final_classification}

To further incorporate the overall deviation from healthy handwriting
patterns, the sample-level normative score is projected into the same
feature space and added to $\mathbf{s}_{\mathrm{risk}}$:
\begin{equation}
    \mathbf{s}^{\prime}_{\mathrm{risk}}
    =
    \mathbf{s}_{\mathrm{risk}}
    +
    \phi_n
    \left(
        \log(1+d_{\mathrm{score}})
    \right),
\end{equation}
where $\phi_n$ is a learnable
projection. The normative score is detached before this operation to
prevent the classification objective from directly optimizing the
healthy normative predictor.

Finally, the AD/HC prediction is
\begin{equation}
    \hat{\mathbf{y}}
    =
    \operatorname{Softmax}
    \left(
        \mathbf{W}_{c}
        \mathbf{s}^{\prime}_{\mathrm{risk}}
        +
        \mathbf{b}_{c}
    \right).
\end{equation}

\subsection{Training Objective}
\label{subsec:training_objective}

The model is trained with a compact objective:
\begin{equation}
    \mathcal{L}
    =
    \mathcal{L}_{CE}
    +
    \lambda_{seg}\mathcal{L}_{seg}
    +
    \lambda_{norm}\mathcal{L}_{norm}.
\end{equation}
where $\lambda_{seg}=0.1$ and $\lambda_{norm}=0.2$. These weights keep the auxiliary segment-level and normative objectives effective without overwhelming the primary classification loss, thereby maintaining a balanced and stable optimization.

The sample-level classification loss is the standard cross-entropy loss:
\begin{equation}
    \mathcal{L}_{CE}
    =
    -\frac{1}{B}
    \sum_{i=1}^{B}
    \log
    \hat{y}_{i,y_i}.
\end{equation}

Since no manual segment-level annotations are available, we use weakly supervised segment-level multiple-instance learning. 
For sample $i$, let $L_i=\sum_{j=1}^{T_f}\bar{m}_{i,j}$ be the number of valid segment tokens. 
The number of selected top-risk tokens is
\begin{equation}
    K_i
    =
    \max
    \left(
    1,
    \left\lceil
    \eta L_i
    \right\rceil
    \right),
\end{equation}
where $\eta$ is the top-$k$ ratio. 
The sample-level logit is
\begin{equation}
    \bar{r}_i
    =
    \frac{1}{K_i}
    \sum_{j\in\mathrm{TopK}_i}
    \tilde{r}_{i,j}.
\end{equation}
The segment MIL loss is
\begin{equation}
    \mathcal{L}_{seg}
    =
    -\frac{1}{B}
    \sum_{i=1}^{B}
    \left[
    y_i\log\sigma(\bar{r}_i)
    +
    (1-y_i)
    \log
    \left(
    1-\sigma(\bar{r}_i)
    \right)
    \right].
\end{equation}
This loss encourages AD samples to contain discriminative high-risk segments, while forcing even the most suspicious segments in HC samples to have low AD-like responses.

The healthy normative loss is applied only to healthy control samples:
\begin{equation}
    \mathcal{L}_{norm}
    =
    \frac{
        \sum_{i\in\mathcal{B}_{HC}}
        \sum_{j=1}^{T_f-1}
        m^{\mathrm{step}}_{i,j}
        \frac{1}{F}
        \left\|
        \hat{\bar{\mathbf{x}}}_{i,j+1}
        -
        \bar{\mathbf{x}}_{i,j+1}
        \right\|_2^2
    }{
        \sum_{i\in\mathcal{B}_{HC}}
        \sum_{j=1}^{T_f-1}
        m^{\mathrm{step}}_{i,j}
    },
\end{equation}
where $\mathcal{B}_{HC}$ is the set of healthy control samples in the batch. If no healthy-control sample is present in a mini-batch,
$\mathcal{L}_{norm}$ is set to zero.
AD samples do not contribute to this loss, so the normative
predictor learns healthy segment-level trajectory regularities
rather than disease-specific patterns.

% \subsection{High-Risk Segment Discovery}
% \label{subsec:high_risk_segment_discovery}

% For interpretability, NormPaST-Risk directly selects high-risk segments from the state-consistent segment sequence.
% Each segment \(F_j\) corresponds to a continuous raw-point interval \(I_j=[s_j,e_j)\) and belongs exclusively to either the Paper or Air state. Given the learned segment-risk weights \(\{\alpha_j\}_{j=1}^{T_f}\), the valid segments are ranked by \(\alpha_j\), and the top \(\rho_f\) proportion is retained.
% The number of selected segments is
% \begin{equation}
% K
% =
% \max\left(
% 1,
% \left\lceil
% \rho_f T_f
% \right\rceil
% \right).
% \end{equation}

% Since the raw-point boundaries of each segment are preserved during tokenization, every selected segment can be directly mapped back to its original trajectory interval \(I_j=[s_j,e_j)\).
% This enables direct visualization of localized high-risk handwriting regions while preserving their Paper/Air state identity.

\section{Experiments}

\subsection{Dataset}
\label{subsec:dataset}

We used the DARWIN-RAW dataset~\cite{cilia2022diagnosing,cilia2021online}, a public online handwriting benchmark for Alzheimer's disease (AD) detection. 
The dataset contains recordings from 174 participants, including 89 AD patients and 85 healthy controls (HC). 
Each participant completed 25 handwriting tasks designed for early AD detection, covering memory and dictation (M), copying (C), and graphic (G) tasks. The handwriting data were collected using a Wacom Bamboo tablet at a sampling rate of 200 Hz. 
In our experiments, each participant-task recording is treated as one trajectory sample with a subject-level AD/HC label. 
From the raw signals, we derive six dynamic kinematic features, including velocity, acceleration, jerk, curvature, pressure variation, and angular velocity, which are used as the input sequence of our model.

\subsection{Experimental Setup}
\label{subsec:experimental_setup}

We evaluated the proposed method using five-fold cross-validation on the DARWIN-RAW dataset. 
In each fold, all samples from the same participant were assigned to the same split to avoid subject information leakage. 
Each participant-task recording was treated as one trajectory sample, and all 25 handwriting tasks were used for joint training and evaluation.

For a fair comparison, all baseline models were evaluated under the same subject-level splits, input features, and evaluation protocol as the proposed method. 
We compared our method with classical machine learning and deep sequence modeling baselines, Random Forest (RF)\cite{breiman2001random}, CNN-1D(AD)\cite{dao2022detection}, BiLSTM\cite{schuster1997bidirectional}, Reservoir Computing(AD)\cite{mwamsojo2022reservoir}, Transformer\cite{vaswani2017attention}, Mamba\cite{gu2023mamba}, HSDA-MS(AD)\cite{gong2025hybrid}. 
During training, we used the AdamW optimizer with an initial learning rate of $1\times10^{-4}$ and a weight decay of $3\times10^{-4}$. 
The batch size was set to 8, and the maximum number of training epochs was set to 100. 
A cosine annealing learning rate scheduler was adopted, with the minimum learning rate set to 1\% of the initial learning rate. 
Early stopping was applied when the validation performance did not improve for 10 consecutive epochs. The validation set is randomly split from the training data, accounting for 10\% of the full dataset.
All experiments were implemented in PyTorch and conducted on NVIDIA\texttrademark{} GPUs. 

We use balanced accuracy as the primary metric to account for potential class imbalance between AD and HC participants. We further report Precision-AD, F1-AD, sensitivity, and specificity for a comprehensive evaluation. We additionally report the area under the receiver operating characteristic curve (AUC) to evaluate the model's discriminative ability. 

The main architectural configurations of the proposed NormPaST-Risk model are summarized in Table~\ref{tab:model_config}. 

\begin{table}[ht]
\centering
\caption{Architectural settings of the proposed NormPaST-Risk model.}
\label{tab:model_config}
\begin{tabularx}{0.95\linewidth}{>{\raggedright\arraybackslash}X >{\raggedright\arraybackslash}X}
\toprule
\textbf{Parameter} & \textbf{Value} \\
\midrule
Input Dimension & 6 \\
Hidden Dimension & 96 \\
Multi-scale Kernel Sizes & [3, 7] \\
Segment Token Stride & 128 points \\
Number of SSM Layers & 4 \\
SSM State Dimension & 16 \\
Segment-risk Experts & 8 \\
Expert Hidden Dimension & 48 \\
Router Hidden Dimension & 96 \\
State-change Weight ($\lambda_c$) & 0.5 \\
Normative Risk Weight  ($\lambda_n$) & 0.3 \\
Dropout Rate & 0.3 \\
\bottomrule
\end{tabularx}
\end{table}

\subsection{Comparison with Representative Baselines}
\label{subsec:baseline_comparison}

To evaluate the effectiveness of the proposed NormPaST-Risk, we compare it with a diverse set of representative methods, including a classical machine learning model (Random Forest), recurrent and convolutional neural networks (BiLSTM and CNN-1D), reservoir computing, a Transformer-based model, the selective state-space model Mamba, and the handwriting-oriented HSDA-MS. 
All methods are evaluated under the same experimental protocol, and the results are reported as mean $\pm$ standard deviation (\%) across five folds. 
The best result for each metric is highlighted in bold, while the second-best result is underlined.

\begin{table*}[t]
\centering
\caption{Comparison with representative baseline methods for online handwriting-based AD detection. Results are reported as mean $\pm$ standard deviation (\%) over five folds. The best result is shown in bold and the second-best result is underlined.}
\label{tab:main_comparison}
\small
\setlength{\tabcolsep}{7pt}
\renewcommand{\arraystretch}{1.15}
\begin{tabular}{lcccccc}
\toprule
\textbf{Model}
& \textbf{BAcc}
& \textbf{Precision-AD}
& \textbf{F1-AD}
& \textbf{Sensitivity}
& \textbf{Specificity}
& \textbf{AUC} \\
\midrule

CNN-1D
& 74.80 $\pm$ 8.28
& 71.21 $\pm$ 5.92
& 77.78 $\pm$ 9.22
& 86.67 $\pm$ 16.01
& 62.94 $\pm$ 9.67
& 88.39 $\pm$ 9.94 \\

BiLSTM
& 76.90 $\pm$ 5.47
& 77.80 $\pm$ 8.65
& 78.05 $\pm$ 4.08
& 79.67 $\pm$ 9.66
& 74.12 $\pm$ 16.43
& 84.81 $\pm$ 4.08 \\

RF
& 77.88 $\pm$ 10.73
& 78.24 $\pm$ 14.45
& 79.71 $\pm$ 10.00
& 83.99 $\pm$ 14.80
& 71.76 $\pm$ 22.55
& 85.50 $\pm$ 7.19 \\

Reservoir
& 81.16 $\pm$ 5.00
& 76.56 $\pm$ 3.54
& 83.57 $\pm$ 4.99
& 92.09 $\pm$ 7.53
& 70.22 $\pm$ 4.24
& 90.26 $\pm$ 4.07 \\

Transformer
& 88.92 $\pm$ 6.59
& 85.39 $\pm$ 8.25
& 89.96 $\pm$ 6.26
& 95.49 $\pm$ 7.26
& \underline{82.35 $\pm$ 10.19}
& 92.73 $\pm$ 5.72 \\

Mamba
& 88.82 $\pm$ 6.61
& 84.59 $\pm$ 9.31
& \underline{90.43 $\pm$ 5.31}
& \textbf{97.71 $\pm$ 3.13}
& 79.93 $\pm$ 13.45
& 93.62 $\pm$ 4.21 \\

HSDA-MS
& \underline{89.05 $\pm$ 6.08}
& \underline{86.13 $\pm$ 9.72}
& 90.36 $\pm$ 4.56
& \underline{95.75 $\pm$ 2.84}
& \underline{82.35 $\pm$ 14.41}
& \underline{93.88 $\pm$ 4.15} \\

\midrule

\textbf{Ours}
& \textbf{92.55 $\pm$ 4.34}
& \textbf{92.43 $\pm$ 5.72}
& \textbf{92.74 $\pm$ 4.23}
& 93.33 $\pm$ 6.09
& \textbf{91.77 $\pm$ 6.71}
& \textbf{95.84 $\pm$ 2.90} \\

\bottomrule
\end{tabular}
\end{table*}

As shown in Table~\ref{tab:main_comparison}, NormPaST-Risk achieves the best performance on five of the six evaluation metrics, including an AUC of 95.84\%, a balanced accuracy of 92.55\%, a Precision-AD of 92.43\%, an F1-AD score of 92.74\%, and a specificity of 91.77\%. Compared with the second-best result for each metric, the proposed method improves balanced accuracy, Precision-AD, F1-AD, and specificity by 3.50, 6.30, 2.31, and 9.42 percentage points, respectively, while also achieving a slightly higher AUC.

It is worth noting that the highest sensitivity does not necessarily correspond to the best overall classification performance. Mamba achieves the highest sensitivity of 97.71\%, while its specificity is comparatively lower at 79.93\%. Similarly, Transformer obtains a sensitivity of 95.49\% with a specificity of 82.35\%, suggesting a trade-off between detecting AD participants and correctly identifying healthy controls. This pattern is also reflected in their Precision-AD values of 84.59\% and 85.39\%, respectively, which are not correspondingly high despite their high sensitivity. HSDA-MS shows a similar pattern, achieving 95.75\% sensitivity and 82.35\% specificity, indicating that the performance between the two classes remains somewhat unbalanced.

In contrast, NormPaST-Risk achieves 93.33\% sensitivity together with 91.77\% specificity, resulting in the highest balanced accuracy of 92.55\%. 
This more balanced sensitivity--specificity trade-off suggests that the proposed method can effectively identify AD participants while maintaining reliable recognition of healthy controls. 
Such balanced discrimination is desirable for AD detection, as excessive false-positive predictions may limit the practical reliability of a classification system. These results suggest that relying mainly on a global sequence representation may capture discriminative but less disease-specific variations, whereas identifying localized abnormal handwriting segments can reduce the influence of irrelevant trajectory portions and provide more robust AD-related handwriting evidence.

The performance differences among the sequence models further demonstrate the effectiveness of the proposed architecture. Although BiLSTM, CNN-1D, Reservoir, Transformer, and Mamba are capable of modeling temporal handwriting signals to different extents, their F1-AD scores remain between 77.78\% and 90.43\%, which are lower than the 92.74\% achieved by NormPaST-Risk. Moreover, NormPaST-Risk achieves the highest Precision-AD of 92.43\%, indicating that the proposed model not only identifies AD participants more effectively but also produces fewer false-positive AD predictions. These results suggest that generic temporal modeling alone is insufficient to fully characterize the AD-related abnormalities embedded in online handwriting trajectories.

A key advantage of NormPaST-Risk is that it incorporates handwriting-specific structural priors rather than treating the trajectory as a homogeneous sequence. First, the multi-scale temporal encoder captures handwriting dynamics at different temporal resolutions, enabling the model to characterize both short-term kinematic variations and longer motion patterns. This provides richer local representations than conventional CNN-1D or generic sequence models relying on a single temporal modeling mechanism.

Second, NormPaST-Risk explicitly distinguishes on-paper and in-air behaviors through Paper--Air state-aware modeling. Compared with a generic sequence model, we explicitly introduce some handwriting-specific structural information into the architecture. These two states characterize different aspects of the handwriting process: on-paper trajectories primarily reflect writing execution, whereas in-air movements are associated with spatial transitions and movement preparation. By preserving these heterogeneous states rather than modeling them as an undifferentiated sequence, the proposed framework maintains state-specific dynamics and captures variations across writing execution and transitional behaviors. This provides a more handwriting-specific representation than generic sequence architectures such as BiLSTM, Transformer, and Mamba.

Third, the selective state-space encoder complements local multi-scale representations by capturing long-range dependencies across the handwriting sequence. Rather than relying exclusively on local convolution or recurrent accumulation, it enables interactions between temporally distant segments and characterizes changes in handwriting dynamics over longer temporal ranges. This is particularly useful when informative abnormalities occur intermittently rather than continuously throughout a trajectory. In addition, the Healthy Normative Branch provides an explicit reference for normal handwriting dynamics. Rather than relying solely on direct AD--HC discrimination, it models regular trajectory patterns from HC samples and quantifies deviations from the learned healthy dynamics. This normative deviation provides complementary information to the discriminative representation and encourages the model to focus on handwriting patterns that deviate from normal progression rather than relying only on subject or task variation. 

More importantly, NormPaST-Risk does not rely solely on a global trajectory representation for classification. The multi-expert segment-risk module estimates AD-related predictive evidence at the segment level and emphasizes locally informative regions. By incorporating normative deviation into segment-risk calibration, the model prioritizes discriminative segments while accounting for their deviation from learned healthy dynamics, thereby reducing the influence of less informative trajectory portions.

Overall, NormPaST-Risk combines multi-scale dynamics, Paper-Air state modeling, healthy normative deviation, and segment-level risk discovery to capture handwriting-specific AD evidence more effectively than generic temporal models, resulting in more discriminative and balanced participant-level predictions.

\subsection{Ablation Study}
\label{sec:ablation}

To investigate the contribution of each component in NormPaST-Risk, we conduct a series of ablation experiments under the same five-fold evaluation protocol. For each variant, only the corresponding component is removed or replaced, while the remaining network architecture and training settings are kept unchanged. The ablation results in Table~\ref{tab:ablation} demonstrate that all major components contribute to the overall performance of NormPaST-Risk. Removing the segment-risk module causes the largest degradation in BAcc and F1-AD, decreasing them from 92.55\% and 92.74\% to 88.56\% and 88.46\%, respectively. Sensitivity also drops from 93.33\% to 87.71\%. These results confirm that explicitly identifying and aggregating disease-relevant local segments is more effective than treating all trajectory regions equally.

\begin{table}[t]
\centering
\caption{Ablation study of the proposed NormPaST-Risk model.}
\label{tab:ablation}
\small
\setlength{\tabcolsep}{7pt}
\renewcommand{\arraystretch}{1.15}
\resizebox{\columnwidth}{!}{%
\begin{tabular}{lcccc}
\toprule
\textbf{Variant}
& \textbf{BAcc}
& \textbf{F1-AD}
& \textbf{Sens.}
& \textbf{Spec.}
\\
\midrule
w/o Segment-Risk
& 88.56
& 88.46
& 87.71
& 89.41
 \\
w/o Paper-Air Modulation
& 89.67
& 89.59
& 88.76
& 90.59
 \\
w/o Selective SSM
& 89.74
& 89.33
& 85.36
& \textbf{94.12}
 \\
w/o Multi-scale Encoder
& 90.23
& 90.46
& 91.04
& 89.41
 \\
w/o Healthy Normative Branch
& 90.95
& 90.10
& 89.93
& 91.97
 \\
\midrule
\textbf{Full NormPaST-Risk}
& \textbf{92.55}
& \textbf{92.74}
& \textbf{93.33}
& 91.77
 \\
\bottomrule
\end{tabular}%
}
\end{table}

Removing Paper-Air modulation reduces BAcc to 89.67\% and sensitivity to 88.76\%, indicating that explicitly distinguishing on-paper motor execution from in-air planning behavior provides useful state-dependent information for AD recognition. Similarly, removing the Selective SSM decreases BAcc to 89.74\% and produces the largest reduction in sensitivity, from 93.33\% to 85.36\%, despite achieving a higher specificity of 94.12\%. This suggests that long-range state-space modeling is particularly important for identifying AD participants, while its removal biases the model toward more conservative HC predictions.

The multi-scale encoder also provides consistent improvements across all evaluation metrics. Without multi-scale temporal modeling, BAcc and F1-AD decrease to 90.23\% and 90.46\%, respectively, demonstrating the benefit of capturing handwriting dynamics at different temporal scales. Finally, removing the Healthy Normative Branch reduces BAcc from 92.55\% to 90.95\% and sensitivity from 93.33\% to 89.93\%. Although this degradation is smaller than that caused by removing the other major components, it confirms that deviations from healthy handwriting dynamics provide complementary information for calibrating segment-level risk. 

\begin{figure}[t]
\centering

\subfloat[]{
    \includegraphics[width=0.47\columnwidth]{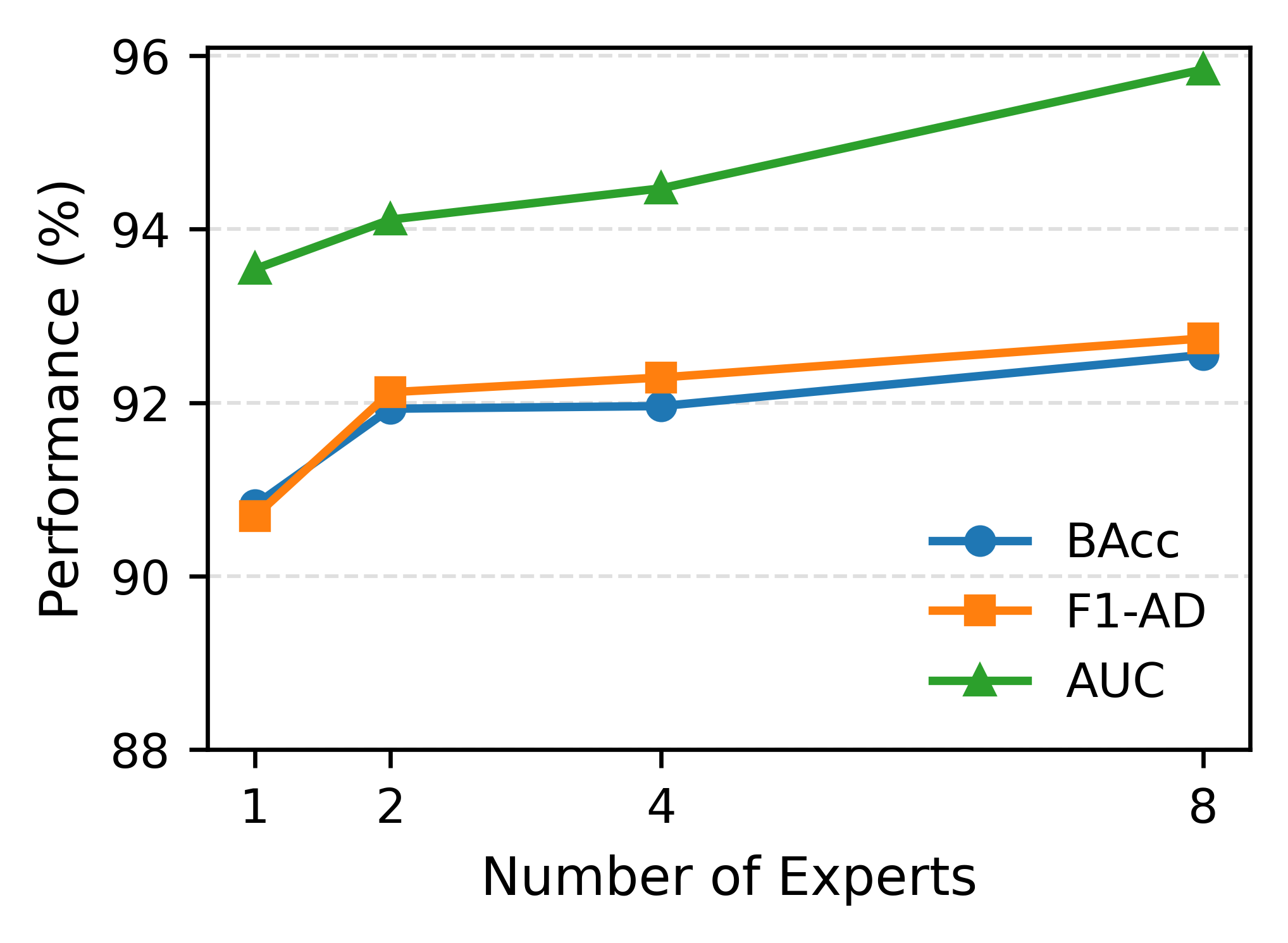}
    \label{fig:num_experts}
}
\hfill
\subfloat[]{
    \includegraphics[width=0.47\columnwidth]{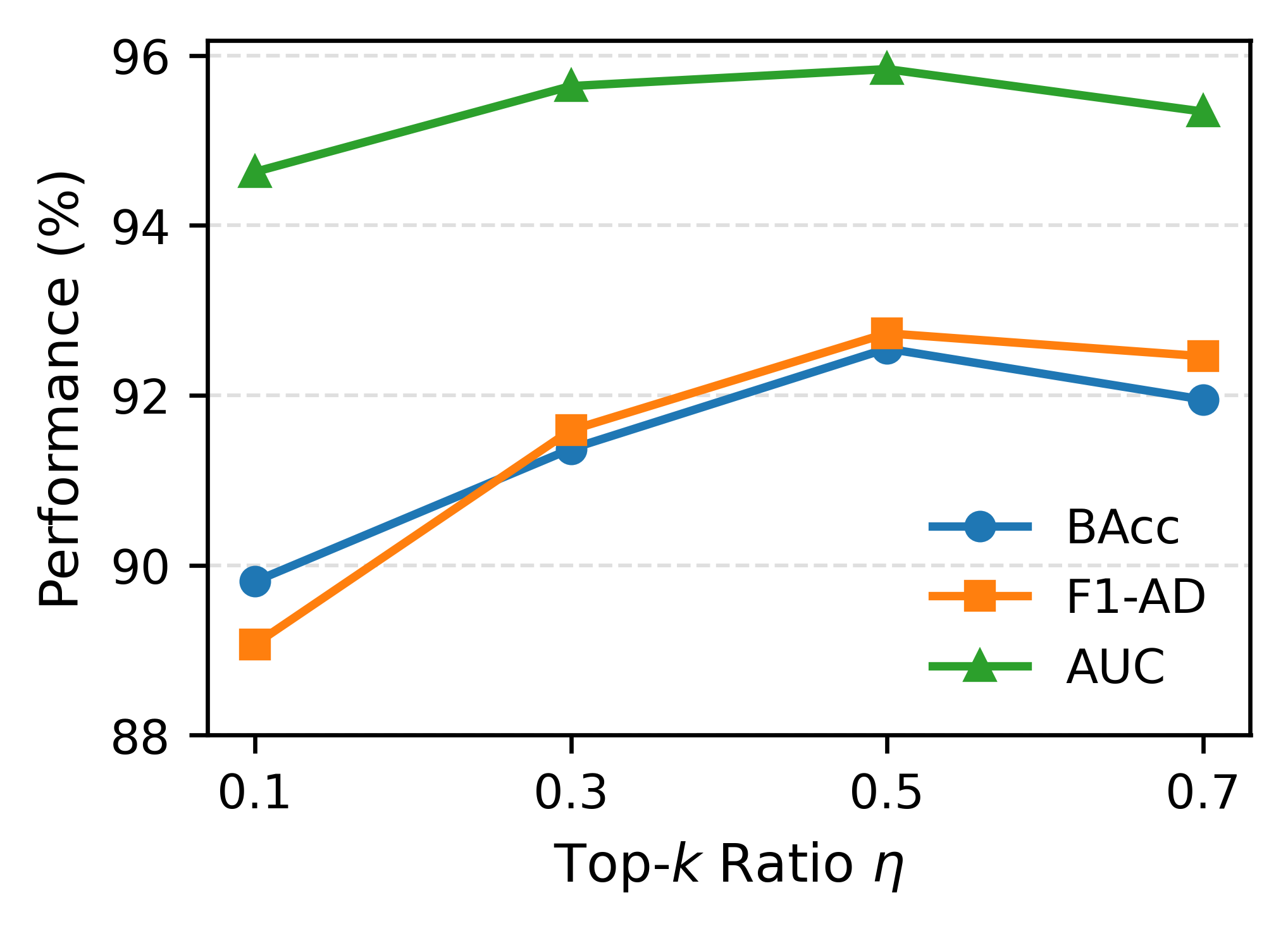}
    \label{fig:topk_ratio}
}

\caption{Performance sensitivity: (a) the number of experts and (b) the top-$k$ ratio $\eta$.}
\label{fig:sensitivity_analysis}
\end{figure}

\begin{table}[t]
\centering
\caption{Participant-level segment-risk statistics for HC and AD groups.}
\label{tab:risk_distribution}
\small
\setlength{\tabcolsep}{3.5pt}
\renewcommand{\arraystretch}{1.15}

\begin{tabular}{lcccc}
\toprule
Aggregation 
& HC 
& AD 
& $\Delta$Risk 
& Cohen's $d$ \\
\midrule
Mean 
& 0.377 $\pm$ 0.014
& 0.435 $\pm$ 0.036
& 0.058
& 2.07 \\

Max 
& 0.466 $\pm$ 0.024
& 0.561 $\pm$ 0.055
& 0.096
& 2.24 \\

Top 50\%
& 0.411 $\pm$ 0.017
& 0.485 $\pm$ 0.046
& 0.074
& 2.12 \\
\bottomrule
\end{tabular}
\end{table}

\begin{table}[t]
\centering
\caption{Performance under segment-level deletion. $\Delta$ denotes the performance decrease relative to the original prediction.}

\label{tab:segment_deletion}
\small
\setlength{\tabcolsep}{6pt}
\renewcommand{\arraystretch}{0.9}

\begin{tabular}{lccccc}
\toprule
Strategy & Ratio & BAcc (\%) & $\Delta$BAcc & AUC (\%) & $\Delta$AUC \\
\midrule
Original & -- & 92.55 & -- & 95.84 & -- \\
\midrule
Low-risk & 0.1 & 91.34 & 1.21 & 95.77 & 0.07 \\
Random   & 0.1 & 91.01 & 1.54 & 95.75 & 0.09 \\
Top-risk & 0.1 & 85.20 & \textbf{7.35} & 93.88 & \textbf{1.96} \\
\midrule
Low-risk & 0.2 & 90.75 & 1.80 & 95.78 & 0.06 \\
Random   & 0.2 & 90.67 & 1.88 & 95.65 & 0.19 \\
Top-risk & 0.2 & 83.01 & \textbf{9.54} & 92.36 & \textbf{3.48} \\
\midrule
Low-risk & 0.3 & 90.36 & 2.19 & 95.51 & 0.33 \\
Random   & 0.3 & 90.37 & 2.18 & 95.35 & 0.49 \\
Top-risk & 0.3 & 83.01 & \textbf{9.54} & 91.98 & \textbf{3.86} \\
\midrule
Low-risk & 0.4 & 90.03 & 2.52 & 95.45 & 0.39 \\
Random   & 0.4 & 89.66 & 2.89 & 95.34 & 0.50 \\
Top-risk & 0.4 & 82.45 & \textbf{10.10} & 91.30 & \textbf{4.54} \\
\midrule
Low-risk & 0.5 & 89.44 & 3.11 & 95.39 & 0.45 \\
Random   & 0.5 & 89.37 & 3.18 & 95.03 & 0.81 \\
Top-risk & 0.5 & 79.71 & \textbf{12.84} & 89.62 & \textbf{6.22} \\
\bottomrule
\end{tabular}
\end{table}

To further examine the robustness of NormPaST-Risk to key design choices in the segment-risk module, we conduct parameter sensitivity analyses on the number of risk experts and the segment top-$k$ ratio. The former determines the capacity of the multi-expert module to model heterogeneous abnormal handwriting patterns, while the latter controls the proportion of high-risk segments selected for weakly supervised segment-risk learning.

Increasing the number of experts generally improves the performance of the segment-risk module. Compared with a single expert, using multiple experts provides clear gains in balanced accuracy, F1-AD, and AUC, suggesting that different experts can capture complementary AD-related handwriting patterns. As shown in Fig.\ref{fig:sensitivity_analysis} (a), the 8-expert configuration achieves the best overall performance. In contrast, the improvement from four to eight experts is relatively moderate, indicating that increasing model capacity beyond a certain point yields diminishing returns. These results support the use of multiple risk experts for modeling heterogeneous AD-related handwriting abnormalities.

The segment top-$k$ ratio controls the proportion of high-risk segments selected for weakly supervised segment-level learning. As shown in Fig.\ref{fig:sensitivity_analysis} (b), as the ratio increases from 0.1 to 0.5, the performance consistently improves, indicating that AD-related abnormalities are distributed across multiple informative handwriting segments rather than being restricted to only a few isolated regions. The performance slightly decreases at 0.7, suggesting that including more lower-ranked segments may dilute discriminative AD-related evidence. A moderate top-$k$ ratio therefore helps retain sufficient abnormal segments while suppressing less informative regions that may contain variations associated with individual writing style, task-specific differences, or acquisition noise. This selective aggregation enables the model to focus more consistently on disease-relevant local patterns rather than irrelevant trajectory variability.

\subsection{Interpretability and Faithfulness Analysis}
\label{sec:segment_faithfulness}

To examine whether the learned segment-risk scores differentiate HC and AD participants, we compare participant-level calibrated risks using mean, maximum, and top-50\% aggregation, as shown in Table~\ref{tab:risk_distribution}. AD participants consistently exhibit higher risk scores than HC participants across all three aggregation strategies. The absolute differences between the two groups are 0.058, 0.096, and 0.074 for mean, maximum, and top-50\% aggregation, respectively.

Although these absolute differences are numerically modest on the $[0,1]$ risk scale, the corresponding Cohen's $d$ values are all greater than 2.0, indicating large standardized differences between the HC and AD groups. Cohen's $d$ was used to quantify the effect size, with values of 0.2, 0.5, and 0.8 conventionally interpreted as small, medium, and large effects, respectively \cite{cohen2013statistical}. In particular, maximum and top-50\% aggregation yield larger group differences than mean aggregation, suggesting that emphasizing high-risk segments enhances the distinction between HC and AD participants. These findings support the segment-level risk modeling strategy and suggest that AD-related predictive evidence is more strongly concentrated in selected high-risk segments rather than being uniformly distributed throughout the trajectory. To further assess the contribution of these high-risk segments to the final prediction, we subsequently perform a segment-level deletion analysis.

To evaluate the faithfulness of the learned segment-risk scores, we perform a post-hoc deletion experiment without model retraining. 
At inference time, valid segments are ranked by their risk scores, and the highest-risk, lowest-risk, or randomly selected segments are masked from segment-risk pooling at ratios $\rho_{\mathrm{del}}\in{0.1,0.2,0.3,0.4,0.5}$.
Random deletion is repeated and averaged, while all model parameters and evaluation settings are kept fixed. 
Results are reported as five-fold participant-level averages.

As shown in Table~\ref{tab:segment_deletion}, removing high-risk segments
consistently causes substantially larger performance degradation than removing
random or low-risk segments. When only 10\% of the highest-risk segments are
removed, BAcc and AUC decrease by 7.35 and 1.96 percentage points,
respectively. In comparison, removing the same proportion of low-risk
segments results in only a 1.21 point decrease in BAcc and almost no change
in AUC. As the deletion ratio increases, the overall degradation becomes more
pronounced. At a deletion ratio of 0.5, removing high-risk segments leads to
decreases of 12.84 percentage points in BAcc and 6.22 percentage points in
AUC, whereas low-risk deletion results in only a 3.11 point decrease in BAcc
and negligible AUC degradation.

These results indicate that the learned segment-risk scores effectively rank
handwriting segments according to their contribution to AD discrimination.
The substantially larger degradation caused by removing high-risk segments
suggests that the localized high-risk regions are closely associated with the
discriminative handwriting evidence used by the model, thereby supporting the faithfulness
of the proposed segment-level risk localization.

To qualitatively assess model interpretability, we visualize one HC participant and one AD participant across the handwriting tasks, while both participants completed all 25 tasks. Similar or visually redundant tasks are omitted for clarity. Each task is presented with three views: the original trajectory (solid lines: on-paper; dashed lines: in-air), the highest-risk segment selected by the model, and the full trajectory risk heatmap.

As shown in Fig.~\ref{fig:normpast}, clear differences can be observed between the two participants across multiple tasks. For the HC participant, most trajectories exhibit low risk, with only sparse regions of slightly elevated risk appearing in more complex tasks. In contrast, the AD participant shows more frequent and pronounced high-risk regions across different task types. Importantly, these regions are localized rather than uniformly distributed over the entire trajectory, indicating that the model relies on specific handwriting portions rather than assigning globally elevated risk to the whole sample.

\begin{strip}
    \centering

    \includegraphics[
        width=0.97\textwidth,
        keepaspectratio
    ]{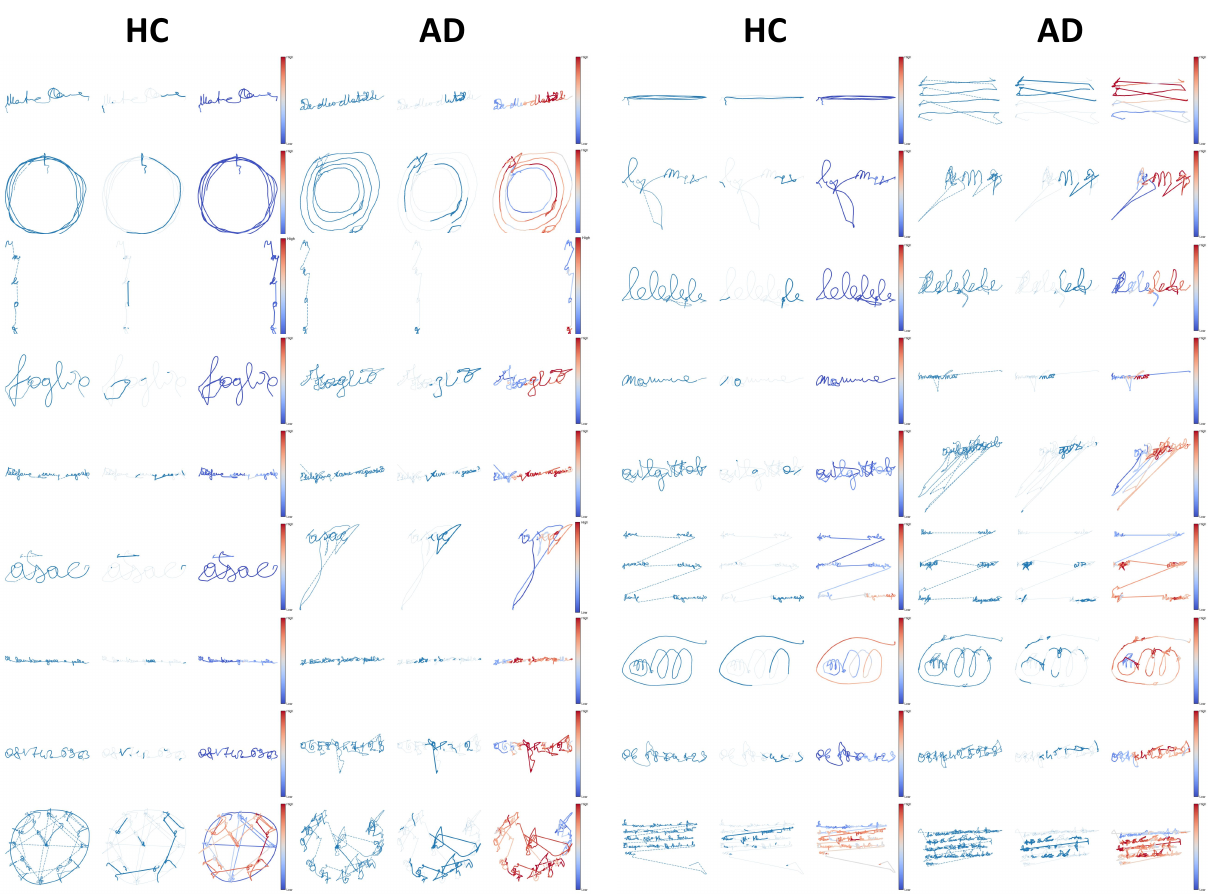}

    \captionof{figure}{
    Qualitative visualization of segment-level risk for one HC participant and one AD participant across selected handwriting tasks. Similar tasks are omitted for clarity. For each task, three subplots are shown: the original trajectory (solid lines indicate on-paper writing and dashed lines indicate in-air movement), the highest-risk segment selected by the model, and the trajectory-level risk heatmap, where warmer colors indicate higher predicted AD-related risk.
    }
    \label{fig:normpast}
\end{strip}

Despite substantial differences in trajectory shape and task content, localized high-risk regions repeatedly appear in the AD participant, whereas the HC participant generally maintains lower risk across the selected tasks. This suggests that the learned segment-risk representation can identify localized disease-related evidence across heterogeneous handwriting tasks, rather than being restricted to a single trajectory pattern.

High-risk regions are often located around portions with relatively complex local dynamics, such as curved strokes, direction changes, stroke transitions, and Paper-Air switching. However, these patterns should be interpreted as model-derived associations rather than direct evidence of specific cognitive or motor impairments. The localized risk distribution further suggests that the model identifies disease-related evidence from selected trajectory segments rather than from uniformly abnormal behavior throughout the sample.

The selected highest-risk segments generally coincide with locally elevated regions in the full risk maps. More importantly, this qualitative observation is consistent with the segment-deletion experiment, where removing high-risk segments results in a substantially larger performance decrease than removing low-risk or randomly selected segments. This provides additional evidence that the identified regions are relevant to the model's AD predictions.

Overall, the visualization suggests that NormPaST-Risk can localize high-risk handwriting segments that contribute to participant-level AD prediction. The contrast between HC and AD risk patterns supports the interpretability of the proposed segment-level risk discovery mechanism. These visualizations should nevertheless be regarded as qualitative model evidence rather than direct clinical biomarkers.

\begin{table}[t]
\centering
\caption{Participant-level kinematic comparisons for segment risk and AD-related high-risk patterns. Values report rank-biserial correlations ($r_{\mathrm{rb}}$) and Benjamini--Hochberg FDR-adjusted $p$-values.}
\label{tab:kinematic_main}

\setlength{\tabcolsep}{3.5pt}
\renewcommand{\arraystretch}{1.15}
\resizebox{\columnwidth}{!}{
\begin{tabular}{lcccc}
\hline
\multirow{2}{*}{Feature}
& \multicolumn{2}{c}{AD: High vs.\ Low}
& \multicolumn{2}{c}{High-risk: AD vs.\ HC}
\\
\cline{2-5}
& $r_{\mathrm{rb}}$
& $p_{\mathrm{FDR}}$
& $r_{\mathrm{rb}}$
& $p_{\mathrm{FDR}}$
\\
\hline

Velocity
& $\mathbf{-0.688}$
& $\mathbf{<0.001}$
& $\mathbf{-0.599}$
& $\mathbf{<0.001}$
\\

Acceleration
& $+0.154$
& $0.208$
& $\mathbf{-0.344}$
& $\mathbf{<0.001}$
\\

Jerk
& $\mathbf{+0.261}$
& $\mathbf{0.039}$
& $+0.016$
& $0.859$
\\

Signed curvature
& $\mathbf{-0.468}$
& $\mathbf{<0.001}$
& $\mathbf{-0.527}$
& $\mathbf{<0.001}$
\\

$d$-pressure
& $\mathbf{-0.967}$
& $\mathbf{<0.001}$
& $+0.137$
& $0.144$
\\

Angular velocity
& $\mathbf{+0.656}$
& $\mathbf{<0.001}$
& $\mathbf{+0.416}$
& $\mathbf{<0.001}$
\\

\hline
\end{tabular}
}
\end{table}

\subsection{Segment-level Kinematic Analysis}
\label{sec:kinematic_analysis}

To further investigate whether the localized segment-risk scores correspond to meaningful AD-related handwriting dynamics, we performed a segment-level kinematic analysis using the six trajectory features. 
For each participant and handwriting task, valid segments were ranked according to their calibrated segment-risk scores, and the top and bottom 20\% were defined as high-risk and low-risk segments, respectively. 
The kinematic features of the selected segments were first averaged within each task and subsequently aggregated at the participant level, such that individual segments were not treated as independent observations. 
High- and low-risk segments were compared using the paired Wilcoxon signed-rank test, whereas comparisons between AD and HC participants were performed using the Mann--Whitney $U$ test. 
Benjamini--Hochberg correction was applied across the six kinematic features, and rank-biserial correlation ($r_{\mathrm{rb}}$) was reported as the effect size.

As summarized in Table~\ref{tab:kinematic_main}, within the AD group, high-risk segments exhibited significantly lower velocity ($r_{\mathrm{rb}}=-0.688$, $p_{\mathrm{FDR}}<0.001$) and lower signed-curvature values ($r_{\mathrm{rb}}=-0.468$, $p_{\mathrm{FDR}}<0.001$), together with higher angular velocity ($r_{\mathrm{rb}}=0.656$, $p_{\mathrm{FDR}}<0.001$), compared with low-risk segments. Pressure variation showed an even stronger association with local risk ($r_{\mathrm{rb}}=-0.967$, $p_{\mathrm{FDR}}<0.001$), whereas acceleration did not differ significantly and jerk showed only a comparatively weaker effect. These results indicate that the learned segment-risk scores are systematically associated with specific local kinematic patterns rather than being arbitrarily distributed along the handwriting trajectory. We next examined whether the localized high-risk segments also exhibited disease-related differences. Compared with HC participants, the high-risk segments of AD participants showed significantly lower velocity ($r_{\mathrm{rb}}=-0.599$, $p_{\mathrm{FDR}}<0.001$) and lower signed-curvature values ($r_{\mathrm{rb}}=-0.527$, $p_{\mathrm{FDR}}<0.001$), together with higher angular velocity ($r_{\mathrm{rb}}=0.416$, $p_{\mathrm{FDR}}<0.001$). Acceleration was also significantly lower in AD high-risk segments ($r_{\mathrm{rb}}=-0.344$, $p_{\mathrm{FDR}}<0.001$), whereas jerk and pressure variation did not show significant between-group differences within the high-risk regions.

Notably, velocity, signed-curvature values, and angular velocity showed significant differences in both the AD high-versus-low comparison and the AD-versus-HC high-risk comparison, with consistent directions of change. These three features therefore represent the principal overlap between kinematic characteristics associated with local risk localization and those exhibiting disease-related differences within high-risk handwriting segments. In contrast, pressure variation was strongly associated with local risk localization but did not show a significant between-group difference within high-risk segments, suggesting that not all kinematic characteristics associated with localized risk necessarily reflect disease-related differences.

Because the proposed framework explicitly preserves Paper and Air states, we further examined whether the observed kinematic differences could be attributed to differences in state composition. High-risk segments contained a significantly larger proportion of Air segments than low-risk segments (49.95\% vs.\ 40.99\%, $p<0.001$), motivating a state-stratified analysis. The complete results are reported in Table~\ref{tab:kinematic_state}.

Within Paper segments, AD high-risk regions showed significantly lower velocity and signed-curvature values and higher angular velocity than AD low-risk regions. The same three characteristics also showed significant disease-related differences within Paper high-risk segments, with lower velocity and signed-curvature values and higher angular velocity in AD than in HC. Pressure variation exhibited a particularly strong association with local risk in the Paper state but did not significantly differentiate AD from HC within high-risk segments.

Within Air segments, AD high-risk regions similarly showed significantly lower velocity and signed-curvature values and higher angular velocity than AD low-risk regions. In the disease-group comparison, velocity, acceleration, and signed-curvature values were significantly lower in AD than in HC high-risk Air segments, whereas angular velocity did not show a significant between-group difference.

Overall, velocity and signed-curvature values showed disease-related differences within high-risk segments in both Paper and Air states, indicating that these differences were not solely driven by variations in Paper/Air segment composition.

\begin{table}[t]
\centering
\caption{State-stratified kinematic comparisons within Paper and Air segments.
Values report rank-biserial correlations ($r_{\mathrm{rb}}$) with
Benjamini--Hochberg FDR-adjusted $p$-values in parentheses.
$d$-pressure was excluded for Air segments because
it was structurally zero.}
\label{tab:kinematic_state}

\setlength{\tabcolsep}{0.5pt}
\renewcommand{\arraystretch}{1.15}
\resizebox{\columnwidth}{!}{
\begin{tabular}{lcccc}
\hline
\multirow{2}{*}{Feature}
& \multicolumn{2}{c}{Paper}
& \multicolumn{2}{c}{Air}
\\
\cline{2-5}
& AD: High vs.\ Low
& High-risk: AD vs.\ HC
& AD: High vs.\ Low
& High-risk: AD vs.\ HC
\\
\hline

Velocity
& $\mathbf{-0.865}\;(\mathbf{<0.001})$
& $\mathbf{-0.526}\;(\mathbf{<0.001})$
& $\mathbf{-0.266}\;(\mathbf{0.049})$
& $\mathbf{-0.576}\;(\mathbf{<0.001})$
\\

Acceleration
& $\mathbf{+0.321}\;(\mathbf{0.009})$
& $-0.124\;(0.192)$
& $-0.008\;(0.946)$
& $\mathbf{-0.400}\;(\mathbf{<0.001})$
\\

Jerk
& $\mathbf{+0.564}\;(\mathbf{<0.001})$
& $+0.175\;(0.070)$
& $-0.178\;(0.182)$
& $-0.065\;(0.457)$
\\

Signed curvature
& $\mathbf{-0.343}\;(\mathbf{0.006})$
& $\mathbf{-0.589}\;(\mathbf{<0.001})$
& $\mathbf{-0.337}\;(\mathbf{0.014})$
& $\mathbf{-0.388}\;(\mathbf{<0.001})$
\\

$d$-pressure
& $\mathbf{-0.970}\;(\mathbf{<0.001})$
& $+0.066\;(0.450)$
& --
& --
\\

Angular velocity
& $\mathbf{+0.582}\;(\mathbf{<0.001})$
& $\mathbf{+0.460}\;(\mathbf{<0.001})$
& $\mathbf{+0.454}\;(\mathbf{<0.001})$
& $+0.143\;(0.129)$
\\
\hline
\end{tabular}
}
\end{table}

\section{Conclusion}

In this work, we proposed NormPaST-Risk, a trajectory-based framework for online handwriting-based AD detection. The model integrates multi-scale temporal encoding, Paper-Air state-aware modeling, selective state-space modeling, healthy normative learning, and task-aware multi-expert segment-risk estimation to capture both global handwriting dynamics and localized AD-related abnormalities. In addition to participant-level prediction, the learned segment risks enable high-risk segments to be projected back onto the original handwriting trajectory, providing interpretable handwriting evidence for AD detection.

Experiments demonstrate that NormPaST-Risk achieves strong and balanced classification performance, reaching 92.55\% BAcc and 95.84\% AUC at the participant level. Ablation studies verify the contributions of the main components, while segment-deletion experiments show that removing high-risk segments causes substantially larger performance degradation than removing low-risk or randomly selected segments, supporting the reliability of the discovered local evidence. Overall, these results suggest that explicitly modeling Paper-Air dynamics and localized abnormal handwriting patterns provides an effective alternative to relying solely on global trajectory representations.

Several limitations remain. First, the current segment construction uses a fixed length of 128 trajectory points. Although this provides a consistent temporal unit for segment-level modeling, the resulting boundaries do not necessarily align with semantic handwriting units such as complete strokes, letters, or words. Consequently, a meaningful writing unit may be split across multiple segments, potentially weakening semantic continuity and limiting the interpretability of the discovered high-risk regions. Future work could explore adaptive or stroke-aware segmentation to better preserve meaningful handwriting structures. Second, the current evaluation is mainly based on a single online handwriting dataset, and the robustness of the proposed framework across different acquisition devices, handwriting tasks, and populations still requires further validation. Third, the segment-risk module is learned under weak supervision without manual segment-level annotations, meaning that the identified high-risk regions should be interpreted as model-derived evidence rather than direct clinical biomarkers. Finally, the healthy normative branch relies on HC samples to learn normal handwriting dynamics, which may be sensitive to the diversity and representativeness of the healthy reference population. Future work will therefore focus on adaptive segmentation, cross-dataset validation, more robust normative modeling, and clinical verification of the discovered high-risk handwriting patterns.

\bibliographystyle{IEEEtran}

% Loading bibliography database
\bibliography{cas-refs}

@article{sweidan2024explainability,
  title={Explainability of CNN-based Alzheimer’s disease detection from online handwriting},
  author={Sweidan, Jana and El-Yacoubi, Mounim A and Rigaud, Anne-Sophie},
  journal={Scientific Reports},
  volume={14},
  number={1},
  pages={22108},
  year={2024},
  publisher={Nature Publishing Group UK London}
}

@article{mwamsojo2022reservoir,
  title={Reservoir computing for early stage Alzheimer’s disease detection},
  author={Mwamsojo, Nickson and Lehmann, Frederic and El-Yacoubi, Mounim A and Merghem, Kamel and Frignac, Yann and Benkelfat, Badr-Eddine and Rigaud, Anne-Sophie},
  journal={IEEE Access},
  volume={10},
  pages={59821--59831},
  year={2022},
  publisher={IEEE}
}

@article{ghaderyan2018new,
  title={A new algorithm for kinematic analysis of handwriting data; towards a reliable handwriting-based tool for early detection of Alzheimer's disease},
  author={Ghaderyan, Peyvand and Abbasi, Ataollah and Saber, Sajad},
  journal={Expert Systems with Applications},
  volume={114},
  pages={428--440},
  year={2018},
  publisher={Elsevier}
}

@article{cilia2024word,
  title={How word semantics and phonology affect handwriting of Alzheimer’s patients: a machine learning based analysis},
  author={Cilia, Nicole D and De Stefano, Claudio and Fontanella, Francesco and Siniscalchi, Sabato Marco},
  journal={Computers in Biology and Medicine},
  volume={169},
  pages={107891},
  year={2024},
  publisher={Elsevier}
}

@article{qi2025alzheimer,
  title={Alzheimer’s disease digital biomarkers multidimensional landscape and AI model scoping review},
  author={Qi, Wenhao and Zhu, Xiaohong and Wang, Bin and Shi, Yankai and Dong, Chaoqun and Shen, Shiying and Li, Jiaqi and Zhang, Kun and He, Yunfan and Zhao, Mengjiao and others},
  journal={npj Digital Medicine},
  volume={8},
  number={1},
  pages={366},
  year={2025},
  publisher={Nature Publishing Group UK London}
}

@article{vcepukaityte2024early,
  title={Early detection of diseases causing dementia using digital navigation and gait measures: A systematic review of evidence},
  author={{\v{C}}epukaityt{\.e}, Giedr{\.e} and Newton, Coco and Chan, Dennis},
  journal={Alzheimer's \& Dementia},
  volume={20},
  number={4},
  pages={3054--3073},
  year={2024},
  publisher={Wiley Online Library}
}

@article{karikari2022blood,
  title={Blood phospho-tau in Alzheimer disease: analysis, interpretation, and clinical utility},
  author={Karikari, Thomas K and Ashton, Nicholas J and Brinkmalm, Gunnar and Brum, Wagner S and Benedet, Andr{\'e}a L and Montoliu-Gaya, Laia and Lantero-Rodriguez, Juan and Pascoal, Tharick Ali and Suarez-Calvet, Marc and Rosa-Neto, Pedro and others},
  journal={Nature Reviews Neurology},
  volume={18},
  number={7},
  pages={400--418},
  year={2022},
  publisher={Nature Publishing Group UK London}
}

@article{jack2024revised,
  title={Revised criteria for diagnosis and staging of Alzheimer's disease: Alzheimer's Association Workgroup},
  author={Jack Jr, Clifford R and Andrews, J Scott and Beach, Thomas G and Buracchio, Teresa and Dunn, Billy and Graf, Ana and Hansson, Oskar and Ho, Carole and Jagust, William and McDade, Eric and others},
  journal={Alzheimer's \& Dementia},
  volume={20},
  number={8},
  pages={5143--5169},
  year={2024},
  publisher={Wiley Online Library}
}

@article{knopman2021alzheimer,
  title={Alzheimer disease},
  author={Knopman, David S and Amieva, Helene and Petersen, Ronald C and Ch{\'e}telat, G{\"a}el and Holtzman, David M and Hyman, Bradley T and Nixon, Ralph A and Jones, David T},
  journal={Nature reviews Disease primers},
  volume={7},
  number={1},
  pages={33},
  year={2021},
  publisher={Nature Publishing Group UK London}
}

@article{kourtis2019digital,
  title={Digital biomarkers for Alzheimer’s disease: the mobile/wearable devices opportunity},
  author={Kourtis, Lampros C and Regele, Oliver B and Wright, Justin M and Jones, Graham B},
  journal={NPJ digital medicine},
  volume={2},
  number={1},
  pages={9},
  year={2019},
  publisher={Nature Publishing Group UK London}
}

@article{tsoi2015cognitive,
  title={Cognitive tests to detect dementia: a systematic review and meta-analysis},
  author={Tsoi, Kelvin KF and Chan, Joyce YC and Hirai, Hoyee W and Wong, Samuel YS and Kwok, Timothy CY},
  journal={JAMA internal medicine},
  volume={175},
  number={9},
  pages={1450--1458},
  year={2015}
}

@article{hampel2018blood,
  title={Blood-based biomarkers for Alzheimer disease: mapping the road to the clinic},
  author={Hampel, Harald and O’Bryant, Sid E and Molinuevo, Jos{\'e} L and Zetterberg, Henrik and Masters, Colin L and Lista, Simone and Kiddle, Steven J and Batrla, Richard and Blennow, Kaj},
  journal={Nature Reviews Neurology},
  volume={14},
  number={11},
  pages={639--652},
  year={2018},
  publisher={Nature Publishing Group UK London}
}

@book{cohen2013statistical,
  title={Statistical power analysis for the behavioral sciences},
  author={Cohen, Jacob},
  year={2013},
  publisher={routledge}
  }

@article{kang2024early,
  title={Early Alzheimer's disease diagnosis via handwriting with self-attention mechanisms},
  author={Kang, Lei and Zhang, Xiaolei and Guan, Jitian and Huang, Kai and Wu, Renhua},
  journal={Journal of Alzheimer’s Disease},
  volume={102},
  number={1},
  pages={173--180},
  year={2024},
  publisher={SAGE Publications Sage UK: London, England}
}

@article{nardone2025handwriting,
  title={Handwriting strokes as biomarkers for Alzheimer’s disease prediction: a novel machine learning approach},
  author={Nardone, Emanuele and De Stefano, Claudio and Cilia, Nicole Dalia and Fontanella, Francesco},
  journal={Computers in Biology and Medicine},
  volume={190},
  pages={110039},
  year={2025},
  publisher={Elsevier}
}

@inproceedings{perez2018film,
  title={Film: Visual reasoning with a general conditioning layer},
  author={Perez, Ethan and Strub, Florian and De Vries, Harm and Dumoulin, Vincent and Courville, Aaron},
  booktitle={Proceedings of the AAAI conference on artificial intelligence},
  volume={32},
  number={1},
  year={2018}
}

@article{muller2017increased,
  title={Increased diagnostic accuracy of digital vs. conventional clock drawing test for discrimination of patients in the early course of Alzheimer’s disease from cognitively healthy individuals},
  author={M{\"u}ller, Stephan and Preische, Oliver and Heymann, Petra and Elbing, Ulrich and Laske, Christoph},
  journal={Frontiers in aging neuroscience},
  volume={9},
  pages={101},
  year={2017},
  publisher={Frontiers Media SA}
}

@article{fernandes2023handwriting,
  title={Handwriting changes in Alzheimer’s disease: a systematic review},
  author={Fernandes, Carina Pereira and Montalvo, Gemma and Caligiuri, Michael and Pertsinakis, Michael and Guimaraes, Joana},
  journal={Journal of Alzheimer’s Disease},
  volume={96},
  number={1},
  pages={1--11},
  year={2023},
  publisher={SAGE Publications Sage UK: London, England}
}

@article{impedovo2018dynamic,
  title={Dynamic handwriting analysis for the assessment of neurodegenerative diseases: a pattern recognition perspective},
  author={Impedovo, Donato and Pirlo, Giuseppe},
  journal={IEEE reviews in biomedical engineering},
  volume={12},
  pages={209--220},
  year={2018},
  publisher={IEEE}
}

@article{schuster1997bidirectional,
  title={Bidirectional recurrent neural networks},
  author={Schuster, Mike and Paliwal, Kuldip K},
  journal={IEEE transactions on Signal Processing},
  volume={45},
  number={11},
  pages={2673--2681},
  year={1997},
  publisher={Ieee}
}

@article{breiman2001random,
  title={Random forests},
  author={Breiman, Leo},
  journal={Machine learning},
  volume={45},
  number={1},
  pages={5--32},
  year={2001},
  publisher={Springer}
}

@article{park2026paano,
  title={Paano: patch-based representation learning for time-series anomaly detection},
  author={Park, Jinju and Kang, Seokho},
  journal={arXiv preprint arXiv:2602.01359},
  year={2026}
}

@inproceedings{liu2026unified,
  title={A unified shape-aware foundation model for time series classification},
  author={Liu, Zhen and Wang, Yucheng and Li, Boyuan and Zheng, Junhao and Eldele, Emadeldeen and Wu, Min and Ma, Qianli},
  booktitle={Proceedings of the AAAI Conference on Artificial Intelligence},
  pages={23972--23980},
  year={2026}
}

@article{gu2021efficiently,
  title={Efficiently modeling long sequences with structured state spaces},
  author={Gu, Albert and Goel, Karan and R{\'e}, Christopher},
  journal={arXiv preprint arXiv:2111.00396},
  year={2021}
}

@article{gu2023mamba,
  title={Mamba: Linear-Time Sequence Modeling with Selective State Spaces},
  author={Gu, Albert and Dao, Tri},
  journal={arXiv preprint arXiv:2312.00752},
  year={2023}
}

@article{gu2022parameterization,
  title={On the parameterization and initialization of diagonal state space models},
  author={Gu, Albert and Goel, Karan and Gupta, Ankit and R{\'e}, Christopher},
  journal={Advances in neural information processing systems},
  volume={35},
  pages={35971--35983},
  year={2022}
}

@article{gupta2022diagonal,
  title={Diagonal state spaces are as effective as structured state spaces},
  author={Gupta, Ankit and Gu, Albert and Berant, Jonathan},
  journal={Advances in neural information processing systems},
  volume={35},
  pages={22982--22994},
  year={2022}
}

@article{gong2025hybrid,
  title={Hybrid Transformer for Early Alzheimer's Detection: Integration of Handwriting-Based 2D Images and 1D Signal Features},
  author={Gong, Changqing and Qin, Huafeng and El-Yacoubi, Moun{\^\i}m A},
  journal={IEEE Journal of Biomedical and Health Informatics},
  year={2025},
  publisher={IEEE}
}

@article{cilia2021online,
  title={From online handwriting to synthetic images for Alzheimer's disease detection using a deep transfer learning approach},
  author={Cilia, Nicole D and D’Alessandro, Tiziana and De Stefano, Claudio and Fontanella, Francesco and Molinara, Mario},
  journal={IEEE Journal of Biomedical and Health Informatics},
  volume={25},
  number={12},
  pages={4243--4254},
  year={2021},
  publisher={IEEE}
}

@article{erdogmus2023promise,
  title={The promise of convolutional neural networks for the early diagnosis of the Alzheimer’s disease},
  author={Erdogmus, Pakize and Kabakus, Abdullah Talha},
  journal={Engineering Applications of Artificial Intelligence},
  volume={123},
  pages={106254},
  year={2023},
  publisher={Elsevier}
}

@article{cilia2022diagnosing,
  title={Diagnosing Alzheimer’s disease from on-line handwriting: A novel dataset and performance benchmarking},
  author={Cilia, Nicole D and De Gregorio, Giuseppe and De Stefano, Claudio and Fontanella, Francesco and Marcelli, Angelo and Parziale, Antonio},
  journal={Engineering Applications of Artificial Intelligence},
  volume={111},
  pages={104822},
  year={2022},
  publisher={Elsevier}
}

@article{dao2022detection,
  title={Detection of Alzheimer disease on online handwriting using 1d convolutional neural network},
  author={Dao, Quang and El-Yacoubi, Moun{\^\i}m A and Rigaud, Anne-Sophie},
  journal={IEEE Access},
  volume={11},
  pages={2148--2155},
  year={2022},
  publisher={IEEE}
}

@article{vaswani2017attention,
  title={Attention is all you need},
  author={Vaswani, Ashish},
  journal={arXiv preprint arXiv:1706.03762},
  year={2017}
}

\end{document}